\documentclass[preprint,twoside,english,sort&compress]{elsarticle}
\usepackage[a4paper,left=30mm,width=150mm,top=25mm,bottom=25mm]{geometry}
\usepackage[T1]{fontenc}
\usepackage[latin9]{inputenc}
\usepackage{amstext}
\makeatletter
\journal{}
\usepackage{xcolor}
\usepackage{amsmath}
\usepackage{amssymb}
\usepackage{slashbox,pict2e} % \slash title for the table
\usepackage{ctex}

\usepackage{Sweave}

\makeatother
\usepackage{amsmath,bm}
\usepackage{algorithm}
\usepackage{algorithmic}
\usepackage{babel}
\usepackage{setspace}
\usepackage{mathrsfs}
\usepackage[algo2e]{algorithm2e}
\def\varneg{\mathord{\raise1.41ex\hbox{\vrule width.4em height.4pt}\kern0pt\vrule width.4pt height1.5ex}}

\newcommand{\BetP}{\mathrm{BetP}}
\graphicspath{{./fig/}}
\begin{document}

\begin{frontmatter}

\title{ECMdd: Evidential $c$-medoids clustering  with multiple prototypes}

%\maketitle
%\iffalse
\author[rvt,focal]{Kuang Zhou\corref{cor1}}

\ead{kzhoumath@163.com}

\author[focal]{Arnaud Martin}

\ead{Arnaud.Martin@univ-rennes1.fr}

\author[rvt]{Quan Pan}
\ead{quanpan@nwpu.edu.cn}
\author[rvt]{Zhun-ga Liu}
\ead{liuzhunga@hotmail.com}
%\ead[url]{http://www.elsevier.com} \fntext[fn1]{This is the specimen author
%footnote.} \fntext[fn2]{Another author footnote, but a little more longer.}
\cortext[cor1]{Corresponding author. Tel.:(+86)029-88431371.}
%\cortext[cor2]{Principal corresponding author}
\address[rvt]{Northwestern Polytechnical University,
Xi'an, Shaanxi 710072, PR China} \address[focal]{DRUID, IRISA, University of Rennes
1, Rue E. Branly, 22300 Lannion, France}
%\fi
\begin{abstract}
%Medoid-based clustering algorithms are of great value for partitioning proximity data sets.
In this work, a new prototype-based clustering method named Evidential $C$-Medoids (ECMdd),  which belongs to the
family of medoid-based clustering for proximity data, is proposed as an extension of Fuzzy $C$-Medoids (FCMdd) on the theoretical framework of belief
functions. In the application of FCMdd and original ECMdd,  a single medoid (prototype), which is supposed to belong to the object set, is utilized to represent one class. For the sake of clarity, this kind of ECMdd using a  single medoid is denoted by sECMdd. In real clustering applications, using only one pattern to capture or interpret a class may not adequately model different types of group structure  and hence limits the clustering performance. In order to  address this problem, a variation of ECMdd using multiple weighted medoids, denoted by wECMdd, is presented.  Unlike sECMdd, in wECMdd objects in each cluster carry various weights  describing their degree of representativeness for that class. This mechanism enables each class to be represented by more than one object. Experimental results in synthetic and real data sets clearly demonstrate the superiority of sECMdd and wECMdd. Moreover, the clustering results by wECMdd can provide richer information for the inner structure of the detected classes with the help of  prototype weights.
\end{abstract}


\end{frontmatter}

\section{Introduction}
Clustering, or unsupervised learning, is a useful technique to detect the underlying cluster structure of the data set. The task of clustering is to
partition  a set of objects $X=\{x_1,x_2,\cdots,x_n\}$ into $c$ groups $\Omega=\{\omega_1,\omega_2,\cdots,\omega_c\}$ in such a way that objects in
the same class are more similar to each other than to those in other classes. The patterns in $X$ are
represented by either object data or relational data. Object data are described explicitly by vectors, while relational data arise from
pairwise similarities or dissimilarities. Among the existing approaches to clustering, the objective function-driven or
prototype-based clustering  such as $C$-Means (CM), Fuzzy $C$-Means (FCM) and Evidential $C$-Means (ECM) is
one of the most widely applied paradigms in statistical pattern recognition.
These methods are based on a fundamentally very simple, but nevertheless very
effective idea, namely to describe the data under consideration by a set of
prototypes. They  capture the characteristics of the data distribution (like
location, size, and shape), and classify the data set based on
the similarities (or dissimilarities) of the objects to their prototypes.

The above mentioned clustering algorithms, CM, FCM and ECM are for object data. The prototype of each class by these methods is the geometrical center of gravity of all the included objects.  But for relational data sets, it is difficult to determine the coordinates of the  centroid of objects. In this case, one of the objects which seems most similar to the ideal center could be set as a prototype. This is the idea of clustering using medoids. Some clustering methods, such as Partitioning Around Medoids (PAM) \cite{kaufman2009finding} and  Fuzzy $C$-Medoids (FCMdd)  \cite{krishnapuram2001low}, produce hard and soft clusters respectively where each of them is represented by a representative medoid. A medoid can be defined as the object of a cluster whose average dissimilarity to all the other objects in the cluster is minimal, {\em i.e.} it is a most centrally located point in the cluster.  However, in real applications, in order to capture various aspects of class structure, it may not
be sufficient enough to use only one object to represent the whole cluster. Consequently we may need more members rather than one to be referred as the prototypes of a group.
%Other than formulating the clustering using medoids, there are some other clustering approaches for relational data in the literature including RFCM, NERFCM

Clustering using multi-prototype has already been studied by some scholars. There are some extensions of FCMdd by using weighted medoids \cite{mei2010fuzzy, mei2011fuzzy} or multiple medoids \cite{de2013relational}. \citet{liu2009multi} proposed a multi-prototype clustering algorithm which can discover the clusters of arbitrary shape and size. In their work,  multiple prototypes
with small separations are organized to model a given number of
clusters in the agglomerative method. New prototypes are iteratively
added to improve the poor cluster boundaries resulted by the poor
initial settings.  \citet{tao2002unsupervised} presented a clustering algorithm adopting multiple centers to represent
the non-spherical shape of classes, and the method could handle non-traditional
curved clusters.  \citet{ghosh2013parameter} considered a multi-prototype classifier which includes options for rejecting
patterns that are ambiguous and/or do not belong to any class. More work about multi-prototype clustering could be found in Refs.~\cite{luo2010multi,ben2011guided}.

Since the boundary between clusters in real-world data sets usually overlaps, soft clustering methods, such as fuzzy clustering, are more suitable than hard clustering for real world applications in data analysis. But the  probabilistic  constraint  of fuzzy memberships  (which must  sum  to  1 across  classes) often brings about some problems, such as  the
inability to distinguish between ``equal evidence" (class membership values
high enough and equal for a number of alternatives) and ``ignorance" (all
class membership values equal but very close to zero)
\citep{menard2000fuzzy,gabrys2000general,guo2015ncm}. Possibility theory and the theory of belief functions \citep{shafer1976mathematical} could been applied to  ameliorate  this problem.

Belief functions have already been  applied in many fields,
such as data classification \cite{denoeux1995k,liu2013new,liu2013evidential,liu2014credal,lian2015evidential,liu2015new,liu2016adaptive}, data clustering \cite{masson2008ecm,denoeux2004evclus, liu2015credal}, social network analysis \cite{wei2013identifying,zhou2014evidential,zhou2015median} and statistical estimation \cite{denoeux2013maximum,come2009learning,zhou2014evidentialem}.  Evidential $C$-Means (ECM) \cite{masson2008ecm} is a newly proposed clustering method to get credal partitions \cite{denoeux2004evclus} for object data. The credal partition is  a
general extension of the  crisp (hard) and fuzzy ones and it allows the object
to belong to not only  single clusters,  but also any subsets of the set of clusters
$\Omega=\{\omega_1,\cdots,\omega_c\}$  by allocating a mass of belief for each object in $X$ over the power
set $2^{\Omega}$. The additional flexibility brought by the power set
provides more refined partitioning results than those by the other  techniques
allowing us to gain a deeper insight into the data \cite{masson2008ecm}. In this paper, we introduce some extensions of FCMdd on the framework of belief functions.  Two versions of evidential $c$-medoids clustering, sECMdd and wECMdd, using a single medoid and multiple weighted medoids respectively to represent a class are proposed to produce the optimal credal partition. The experimental results show the effectiveness of the methods and illustrate the advantages of credal partitions and multi-prototype representation for classes.

The rest of this paper is organized as follows. In Section 2, some basic knowledge and the rationale of our method are briefly introduced. In Section
3 and Section 4, evidential $c$-medoids using a single  medoid and  multiple weighted medoids are presented respectively. Some issues about applying  the algorithms are discussed in Section 5.  In order to show the effectiveness
of the proposed clustering approaches, in Section 6 we test the ECMdd algorithms on different artificial and real-world data sets and make comparisons with related partitive methods. Finally, we conclude and present some perspectives in Section 7.
\section{Background}
In this section some related preliminary  knowledge, including the theory of belief functions and some classical clustering algorithms, will be presented.
\subsection{Theory of belief functions}
Let $\Omega=\{\omega_{1},\omega_{2},\ldots,\omega_{c}\}$ be the finite domain of
$X$, called the discernment frame. The belief functions are defined on the power
set $2^{\Omega}=\{A:A\subseteq\Omega\}$. The function $m:2^{\Omega}\rightarrow[0,1]$ is said to be the Basic Belief
Assignment (bba) on $2^{\Omega}$, if it satisfies:
\begin{equation}
\sum_{A\subseteq\Omega}m(A)=1.
\end{equation}
Every $A\in2^{\Omega}$ such that $m(A)>0$ is called a focal element.
The credibility and plausibility functions are defined as in Eqs.$~\eqref{bel}$ and $\eqref{pl}$ respectively.
%\end{definition}

\begin{equation}
Bel\text{(}A\text{)}=\sum_{B\subseteq A, B \neq \emptyset} m\text{(}B\text{)},~~\forall A\subseteq\Omega,
\label{bel}
\end{equation}

\begin{equation}
 Pl\text{(}A\text{)}=\sum_{B\cap A\neq\emptyset}m\text{(}B\text{)},~~\forall A\subseteq\Omega.
 \label{pl}
\end{equation}
Each quantity $Bel(A)$  measures the total support given to $A$, while $Pl(A)$ represents potential amount of support to $A$. Functions $Bel$ and $Pl$ are linked by the following relation:
\begin{equation}
  Pl(A) = 1 - m(\emptyset)-Bel(\overline{A}),
\end{equation}
where $\overline{A}$ denotes the complement of $A$ in $\Omega$.

A belief function on the credal level can be transformed into a probability function by Smets method \citep{smets2005decision}. In this
algorithm, each mass of belief $m(A)$ is
equally distributed among the elements of $A$.
This leads to the concept of pignistic probability, $\BetP$, defined by
\begin{equation}
 \label{pig}
	\BetP(\omega_i)=\sum_{\omega_i  \in A \subseteq \Omega } \frac{m(A)}{|A|(1-m(\emptyset))},
\end{equation}
where $|A|$ is the number of elements of $\Omega$ in $A$. Pignistic probabilities, which play the same role as fuzzy membership, can
easily help us  make a decision. In fact, belief functions provide us many
decision-making techniques not only  in the form of probability measures. For
instance, a pessimistic decision can be made by maximizing the credibility
function, while maximizing the plausibility function could provide an
optimistic one \citep{martin2008decision}. Another criterion (Appriou's rule)
\citep{martin2008decision} considers the plausibility functions
and consists of attributing the class $A_j$ for  object $i$ if
\begin{equation}
\label{mbx1}
 A_j=\arg \max_{X \subseteq \Omega}
	\{m_i(X)Pl_i(X)\},
\end{equation}
where
\begin{equation} \label{mbx}
	m_i(X)=K_i \lambda_X \left(\frac{1}{|X|^r}\right).
\end{equation}
In Eq.~\eqref{mbx1} $m_i(X)$ is a weight on $Pl_i(X)$, and $r$ is a parameter
in $[0,1]$ allowing a decision from a simple class $(r= 1)$ until the total
ignorance $\Omega$ $(r= 0)$. The value $\lambda_X$ allows the integration of
the lack of knowledge on one of the focal sets $X\subseteq \Omega$, and it can
be set to be 1 simply. Coefficient $K_i$ is the normalization factor to
constrain the mass to be in the closed world:
 \begin{equation}
	K_i=\frac{1}{1-m_i(\emptyset)}.
\end{equation}

\subsection{Evidential $c$-means}
Evidential $c$-means \cite{masson2008ecm} is a direct generalization of FCM in the
framework of belief functions, and it is based on the credal partition first
proposed by \citet{denoeux2004evclus}. The credal partition takes advantage
of imprecise (meta) classes  to express partial knowledge of class memberships.  The principle is different from
another belief clustering method  put forward by
\citet{schubert2004clustering}, in which  conflict
between evidence is utilized to cluster the belief functions related to
multiple events. In ECM, the evidential membership of object
$x_i=\{x_{i1},x_{i2},\cdots,x_{ip}\}$ is represented by a
bba  $\bm{m}_i=\left(m_i\left(A_j\right): A_j \subseteq \Omega \right)$ $(i=1,2,\cdots,n)$ over the given
frame of discernment $\Omega=\{\omega_1,\omega_2,\cdots,\omega_c\}$. The set $\mathcal{F}=\left\{A_j\mid A_j \subseteq \Omega, m_i(A_j)> 0\right\}$ contains all the focal elements. The
optimal credal partition is obtained by  minimizing the following objective
function:
\begin{equation}
	J_{\mathrm{ECM}}=\sum\limits_{i=1}^{n}\sum\limits_{A_j\subseteq \Omega,A_j
	\neq \emptyset}|A_j|^\alpha
	m_{i}(A_j)^{\beta}d_{ij}^2+\sum\limits_{i=1}^{n}\delta^2m_{i}(\emptyset)^{\beta}
	\label{JECM}
\end{equation}
constrained on
\begin{equation}
	\sum\limits_{A_j\subseteq \Omega,A_j \neq
	\emptyset}m_{i}(A_j)+m_{i}(\emptyset)=1, \label{ECMconstraint}
\end{equation}
and
\begin{equation}
\label{nonegcon}
  m_{i}\left(A_j\right) \geq 0, ~~ m_{i}\left(\emptyset \right) \geq 0,
\end{equation}
where $m_{i}(A_j)\triangleq m_{ij}$ is the bba of $x_i$ given
to the nonempty set $A_j$, while $m_{i}(\emptyset)\triangleq m_{i\emptyset}$
is the bba of $x_i$ assigned to the empty set. Parameter $\alpha$ is a tuning parameter allowing to control the
degree of penalization for subsets with high cardinality, parameter $\beta$ is
a weighting exponent and  $\delta$ is an adjustable  threshold  for  detecting
the outliers. Here $d_{ij}$ denotes the
distance (generally Euclidean distance) between $x_i$ and the barycenter ({\em i.e.} prototype,
denoted by $\overline{v}_j$) associated with $A_j$:
\begin{equation}\label{dis_node_pro}
  d_{ij}^2=\|x_i-\overline{v}_j\|^2,
\end{equation}
where $\overline{v}_j$ is defined mathematically by
\begin{equation}\label{prototypes}
	\overline{v}_j=\frac{1}{|A_j|}\sum_{h=1}^c s_{hj} v_h,
	~~\text{with}~~ s_{hj}=
  \begin{cases}
        1 & \text{if} ~~ \omega_h \in A_j, \\
        0 & \text{else}.\end{cases}
 \end{equation}
The notation $v_h$ is the geometrical
center of points in cluster $h$. In fact the value of $d_{ij}$ reflects the distance between object $x_i$ and class $A_j$. Note that a ``noise" class $\emptyset$ is considered in ECM. If $A_j=\emptyset$, it is assumed that the distance between object $x_i$ and  class $A_j$ is $d_{ij}=\delta$. As we can see for credal partitions, the label of class $j$ is not from 1 to $c$
as usual, but ranges in $1,2,\cdots,f$ where $f$ is the number of the focal elements {\em i.e.} $f=|\mathcal{F}|$.
%\begin{equation}
%  f=\left|\{A \subseteq \Omega: m_i(A)>0\}\right|.
%\end{equation}
The update process with Euclidean distance  is given by the
following two alternating steps.
%\iffalse
\begin{enumerate}
\item Assignment update:
\begin{align}
\label{ECM_updata_bba}
m_{ij}=\frac{|A_j|^{-\alpha/(\beta-1)}{d_{ij}^{-2/(\beta-1)}}}{\sum\limits_{A_h\neq\emptyset}
|A_h|^{-\alpha/(\beta-1)}{d_{ih}^{-2/(\beta-1)}}+\delta^{-2/(\beta-1)}},
\forall i, ~\forall j/A_j (\neq \emptyset)\subseteq  \Omega
\end{align}
 \begin{align}
			m_{i\emptyset}=1-\sum_{A_j\neq \emptyset}m_{ij}, ~~\forall
		i=1,2,\cdots,n.
\end{align}
\item Prototype update: The prototypes
		(centers) of the classes are given by the rows of the matrix
		$v_{c\times p}$, which is the solution of the following linear system:
	\begin{equation}
\label{HB}
    \bm{HV}=\bm{B},
   \end{equation}
where $\bm{H}$ is a matrix of size $(c \times c)$ given by
\begin{equation}\label{H}
   \bm{H}_{lt}=\sum_i
			\sum_{A_h \supseteqq \{\omega_t,\omega_l\}} |A_h|^{\alpha-2}
			m_{ih}^\beta, ~~t,l=1,2,\cdots,c,
\end{equation} and $\bm{B}$ is a matrix of size $(c \times p)$ defined by
\begin{equation}\label{B}
			\bm{B}_{lq}=\sum_{i=1}^n x_{iq}\sum_{A_k\ni
			\omega_l}|A_k|^{\alpha-1}m_{ik}^\beta, ~~l=1,2,\cdots,c,~~
			q=1,2,\cdots,p.
\end{equation} \end{enumerate}

\subsection{Hard and fuzzy $c$-medoids clustering}
The hard $C$-Medoids (CMdd) clustering is a variant of the
traditional $c$-means method, and it produces a crisp partition of the data set. Let $\bm{X}=\left\{x_i\mid i=1,2,\cdots,n\right\}$ be the set of $n$ objects and $\tau(x_i, x_j)\triangleq \tau_{ij}$ denote the dissimilarity between objects $x_i$ and  $x_j$. Each object may or may not be represented by a feature vector. Let $\bm{V}= \{v_1, v_2,\cdots,v_c\}$, $v_i \in \bm{X}$ represent a subset of $\bm{X}$. The
objective function of CMdd is similar to that in CM:
\begin{equation}
\label{J_cmdd}
J_\text{CMdd}=\sum_{j=1}^c \sum_{i=1}^n u_{ij}\tau(x_i, v_j),
\end{equation}
where $c$ is the number of clusters. As CMdd is
based on crisp partitions, $u_{ij}$ is either 0 or 1 depending whether
$x_i$ is in cluster $\omega_j$. The notation $v_j$ is the prototype of class $\omega_j$, and it is supposed to be one of the objects in the data set.  Due to the fact that exhaustive search of medoids is an NP hard problem, \citet{kaufman2009finding} proposed one approximate search algorithm called PAM, where the $c$ medoids are found  efficiently. After the selection of the prototypes, object $x_i$ is assigned the closest class $\omega_f$, the medoid of which is most similar to this pattern, {\em i.e.}
\begin{equation}
  x_i \in \omega_f, ~~ \text{with}~~ f = \arg \min_{l = 1,2,\cdots,c} \tau(x_i, v_l).
\end{equation}

Fuzzy $C$-Medoids (FCMdd) is a variation of CMdd designed for relational data \cite{krishnapuram2001low}. The objective function of FCMdd is given as
\begin{equation}
  J_\text{FCMdd} =  \sum_{i=1}^n \sum_{j=1}^c u_{ij}^\beta \tau(x_i, v_j)
\end{equation}
subject to
\begin{equation}
  \sum_{j=1}^c u_{ij}=1, i=1,2,\cdots,n,
\end{equation}
and
\begin{equation}
  u_{ij}\geq 0, i=1,2,\cdots,n, ~~j=1,2,\cdots,c.
\end{equation}
In fact, the objective function of FCMdd is similar to that of FCM. The main difference lies in that the prototype of a class in FCMdd is defined as the medoid, {\em i.e.} one of the object in the original data set, instead of the centroid (the average point in a continuous space) for FCM. The object assignment and prototype selection are preformed by the following alternating update steps:
\begin{enumerate}
\item
		Assignment update:
\begin{equation}
		u_{ij}=\frac{\tau_{ij}^{-1/(\beta-1)}}{\sum\limits_{k=1}^c
	\tau_{ik}^{-1/(\beta-1)}}.
\end{equation}
\item Prototype update: the new
	prototype of cluster $\omega_j$ is set to be $v_{j}=x_{l^*}$ with
	\begin{equation} x_{l^*}= \arg \min_{\{v_j:v_j=x_l
	(\in X)\}} \sum_{i=1}^n u_{ij}^\beta \tau(x_i,v_j).
 \end{equation}
\end{enumerate}
\subsection{Fuzzy clustering with multi-medoid}
In a recent work of \citet{mei2011fuzzy}, a generalized medoid-based Fuzzy clustering with Multiple Medoids (FMMdd) has been proposed. For a data set $\bm{X}$ given the dissimilarity matrix $\bm{R}=\{r_{ij}\}_{n\times n}$, where $r_{ij}$ records the dissimilarity between each two objects $x_i$ and $x_j$. The objective of FMMdd is to minimize the following criterion:
\begin{equation}
  J_\text{FMMdd} = \sum_{k=1}^c \sum_{i=1}^n \sum_{j=1}^n u_{ik}^\beta v_{kj}^\psi r_{ij}
\end{equation}
subject to
\begin{equation}
  \sum_{k=1}^c u_{ik}=1, \forall i=1,2,\cdots n;~~ u_{ik}\geq 0, \forall i ~\text{and}~ k
\end{equation}
and
\begin{equation}
  \sum_{j=1}^n v_{kj}=1, \forall k=1,2,\cdots,c; ~~ v_{kj}\geq 0,~~ \forall k ~\text{and}~ j,
\end{equation}
where $u_{ik}$ denotes the fuzzy membership of $x_i$ for cluster $\omega_k$, and $v_{kj}$ denotes the prototype weights of $x_j$ for cluster $\omega_k$. The constrained minimization problem of finding the optimal fuzzy partition could be solved by the use of Lagrange multipliers and the update equations of $u_{ik}$ and $v_{kj}$ are derived as below:
\begin{equation}
\label{fmmddu}
  u_{ik}= \frac{\left(\sum\limits_{j=1}^n v_{kj}^\psi r_{ij}\right)^{-1/(\beta-1)}}{\sum\limits_{f=k}^c\left(\sum\limits_{j=1}^n v_{fj}^\psi r_{ij}\right)^{-1/(\beta-1)}}
\end{equation}
and
\begin{equation}
\label{fmmddv}
  v_{kj}=\frac{\left(\sum\limits_{i=1}^n u_{ik}^\beta r_{ij}\right)^{-1/(\psi-1)}}{\sum\limits_{h=1}^n\left(\sum\limits_{i=1}^n u_{ik}^\beta r_{ih}\right)^{-1/(\psi-1)}}.
\end{equation}
The FMMdd algorithm starts with a non-negative initialization, then the membership values and prototype weights are iteratively updated with Eqs.~\eqref{fmmddu} and \eqref{fmmddv} until convergence.

%\subsection{Other dissimilarity-based clustering}

\section{sECMdd with a single medoid}
We start with the introduction of  evidential $c$-medoids clustering algorithm using a single medoid, sECMdd,  in order to take advantages of both medoid-based clustering and credal partitions. This partitioning evidential clustering algorithm is mainly related to the fuzzy $c$-medoids.  Like all the prototype-based clustering methods, for sECMdd,
an objective function should first be  found to provide an immediate measure of
the quality of the partitions. Hence our goal can  be characterized as
the optimization of the objective function to get the best credal partition.
\subsection{The objective function}
As before, let $\bm{X}=\left\{x_i\mid i=1,2,\cdots,n\right\}$ be the set of $n$ objects and $\tau(x_i, x_j)\triangleq \tau_{ij}$ denote the dissimilarity between objects $x_i$ and  $x_j$. The pairwise dissimilarity is the only information required for the analyzed data set.
%Each object may or may not be represented by a feature vector.
The objective function of sECMdd is similar to that in ECM:
\begin{equation} J_{\mathrm{sECMdd}}(\bm{M},
	\bm{V})=\sum\limits_{i=1}^{n}\sum\limits_{A_j\subseteq \Omega,A_j \neq
	\emptyset}|A_j|^\alpha
	m_{ij}^{\beta}d_{ij}+\sum\limits_{i=1}^{n}\delta^2m_{i\emptyset}^{\beta},
	\label{costfun} \end{equation}
\noindent constrained on
\begin{equation}
	\sum\limits_{A_j\subseteq \Omega,A_j \neq
	\emptyset}m_{ij}+m_{i\emptyset}=1, \label{sECMddconstraint}
\end{equation}
where $m_{ij}\triangleq m_{i}(A_j)$ is the bba of $x_i$ given to the nonempty
set $A_j$, $m_{i\emptyset} \triangleq m_{i}(\emptyset)$ is the bba of $x_i$
assigned to the empty set, and $d_{ij}\triangleq d(x_i, A_j)$ is the dissimilarity
between $x_i$ and  focal set $A_j$. Parameters $\alpha,\beta,\delta$ are adjustable with the same meanings
as those in ECM. Note that $J_\text{sECMdd}$ depends on the credal partition $\bm{M}$ and the set $\bm{V}$ of all
prototypes.

Let  $v_k^\Omega$ be the prototype of  specific  cluster (whose focal element is a singleton) $A_j=\{\omega_k\}$ $(k=1,2,\cdots,c)$ and assume that it must be one of the objects in $X$. The dissimilarity between object $x_i$ and cluster (focal set) $A_j$  can be defined as follows. If $\left|A_j\right|=1$, {\em i.e.} $A_j$ is associated with one of the
singleton clusters in $\Omega$ (suppose to be $\omega_k$ with prototype
$v_k^\Omega$, {\em i.e.} $A_j=\{\omega_k\}$), then the dissimilarity between $x_i$ and $A_j$ is defined by
\begin{equation} \label{specific_pro}
	d_{ij}=d(x_i, A_j) = \tau(x_i,v_k^\Omega). \end{equation}
When $|A_j|>1$, it represents an imprecise (meta)
cluster. If object $x_i$ is to be partitioned into a meta cluster, two
conditions should be satisfied \cite{zhou2015median}. One condition is  the dissimilarity values between
$x_i$ and the included singleton classes'  prototypes are small. The
other condition is the object should be close to   the prototypes of all these  specific
clusters.  The former measures the degree of  uncertainty, while the latter is
to avoid the pitfall of  partitioning two data objects irrelevant to any
included specific clusters into the corresponding imprecise classes.  Therefore, the medoid (prototype) of an imprecise class $A_j$ could be set to be one of the objects locating with similar dissimilarities to all the prototypes of the specific classes $\omega_k \in A_j$ included in $A_j$. The variance of the dissimilarities of object $x_i$ to the medoids of all the involved specific classes could be taken into account to express the degree of uncertainty. The smaller the variance is, the higher uncertainty we have for object $x_i$. Meanwhile the medoid should be close to all the prototypes of the specific
classes.  This is to distinguish the outliers, which may have similar dissimilarities to the prototypes of some specific classes, but obviously not a good choice for representing the associated imprecise classes. Let $v^{2^\Omega}_j$ denote the medoid of class $A_j$\footnote{The notation $v^\Omega_k$ denotes the prototype of specific class $\omega_k$, indicating it is in the framework of $\Omega$. Similarly, $v^{2^\Omega}_j$ is defined on the power set $2^\Omega$, representing the prototype of the focal set $A_j$ $\in 2^\Omega$. In fact $\bm{V}$ is the set of all the prototypes, {\em i.e.} $\bm{V}=\{v_j^{2^\Omega}:j=1,2,\cdots,2^c-1\}$. It is easy to see
$\{v_k^{\Omega}:k=1,2,\cdots,c\}\subseteq \bm{V} \subseteq \bm{X}$.}. Based on the above analysis, the medoid of $A_j$ should set to $v_j^{2^\Omega}=x_p$ with
\begin{align}
\label{imepro}
  p = \arg \min\limits_{i:x_i\in \bm{X}} \Big\{ f\left(\{\tau(x_i, v_k^\Omega); \omega_k \in A_j \}\right) + \eta \frac{1}{|A_j|} \sum\limits_{\omega_k \in A_j} \tau(x_i, v_k^\Omega)\Big\},
\end{align}
where $\omega_k$ is the element of $A_j$, $v_k^\Omega$ is its corresponding prototype and $f$ denotes the function describing the variance among the related dissimilarity values. The  variance function could be used directly:
\begin{equation}
  \text{Var}_{ij}= \frac{1}{|A_j|}\sum_{\omega_k \in A_j} \bigg[\tau(x_i, v_k^{\Omega})-\frac{1}{|A_j|}\sum_{\omega_k \in A_j}\tau(x_i, v_k^{\Omega})\bigg]^2.
\end{equation}
In this paper, we use the following function to describe  the variance $\rho_{ij}$ of the dissimilarities between object $x_i$ and the medoids of the involved specific classes in $A_j$
\begin{equation}
  \rho_{ij}= \frac{1}{\text{choose}(|A_j|,2)}\sum\limits_{\omega_x, \omega_y \in A_j} \sqrt{\left(\tau(x_i, v_x^{\Omega})-\tau(x_i, v_y^{\Omega})\right)^2},
\end{equation}
where  $\text{choose}(a,b)$ is the number of combinations of the given $a$ elements taken $b$ at a time. Then the dissimilarity between objects $x_i$ and class $A_j$ can be defined as
\begin{equation}
  d_{ij} = \frac{\tau(x_i, v^{2^\Omega}_j) + \gamma \frac{1}{|A_j|} \sum\limits_{\omega_k \in A_j} \tau(x_i, v_k^\Omega)}{1+\gamma}.
\end{equation}
As we can see from the above equation, the dissimilarity between object $x_i$ and meta class $A_j$  is the weighted average of dissimilarities of $x_i$ to the all involved  singleton cluster medoids and to the prototype of the imprecise class  $A_j$ with a tuning factor $\gamma$. If $A_j$ is a specific class with $A_j=\{\omega_k\}$ ($|A_j|=1$), the dissimilarity between $x_j$ and $A_j$ degrades to the dissimilarity between $x_i$ and $v_k^{\Omega}$ as defined in Eq.~\eqref{specific_pro}, {\em i.e.} $v^{2^\Omega}_j=v_k^\Omega$. And if $|A_j|>1$, its medoid is determined by Eq.~\eqref{imepro}.

\noindent\textbf{Remark 1:}  sECMdd is similar to Median Evidential $C$-Means (MECM) \cite{zhou2015median} algorithm.  MECM is in the framework of median clustering, while sECMdd consists with FCMdd in principle. Another difference of sECMdd and MECM is the way of calculating the dissimilarities between objects and imprecise classes.   Although both MECM and sECMdd consider the dissimilarities of objects to the prototypes for specific clusters, the strategy adopted by sECMdd is more simple and intuitive, hence makes sECMdd run faster in real time. Moreover, there is no representative medoid for imprecise classes in MECM.
\subsection{The optimization}
To minimize $J_\text{sECMdd}$, an optimization
scheme via an Expectation-Maximization (EM) algorithm  can be
designed, and the alternating update steps are as follows:

\noindent Step 1. Credal partition ($\bm{M}$) update.

The bbas of objects' class membership for any subset $A_j \subseteq \Omega$ and the empty set $\emptyset$ representing the outliers  are updated identically to ECM \cite{masson2008ecm}:
\begin{enumerate} \item
			$\forall A_j \subseteq \Omega, A_j \neq \emptyset$,
			\begin{equation} \label{mass1}
m_{ij}=\frac{|A_j|^{-\alpha/(\beta-1)}{d_{ij}^{-1/(\beta-1)}}}
{\sum\limits_{A_k\neq\emptyset}|A_k|^{-\alpha/(\beta-1)}{d_{ik}^{-1/(\beta-1)}}+\delta^{-1/(\beta-1)}}
		\end{equation}
\item If $A_j = \emptyset$, \begin{equation}
				\label{mass2} m_{i\emptyset}=1-\sum\limits_{A_j \neq
				\emptyset}m_{ij} \end{equation} \end{enumerate} \noindent Step
2. Prototype ($\bm{V}$) update.

The prototype $v_i^\Omega$ of a specific (singleton) cluster $\omega_i$
$(i=1,2,\cdots,c)$ can be updated first and then the prototypes of imprecise (meta) classes could be determined by Eq.~\eqref{imepro}.  For
singleton clusters $\omega_k$ $(k=1,2,\cdots,c)$, the corresponding new
prototype $v_k^\Omega$ $(k=1,2,\cdots,c)$  could be set to $x_l \in \bm{X}$ such that
  \begin{equation}
\label{pro_update}
 x_l= \arg \min_{v^{'}_k}
	\left\{\sum_{i=1}^n  \sum_{A_j = \{\omega_k\}}
	 m_{ij}^\beta d_{ij}(v^{'}_k): v^{'}_k
	\in X\right\}.
\end{equation}
The dissimilarity between object $x_i$ and cluster $A_j$, $d_{ij}$, is a function of $v^{'}_k$, which is the potential prototype of class $\omega_k$.
%\end{enumerate}

The bbas of the objects' class assignment are updated  identically to ECM
\cite{masson2008ecm}, but it is worth noting that $d_{ij}$ has a
different meaning as that in ECM although in both cases it measures the dissimilarity between object $x_i$ and class $A_j$. In ECM $d_{ij}$ is the distance between object $i$ and the centroid point of $A_j$, while in sECMdd, it is the dissimilarity between $x_i$ and the most ``possible" medoid.  For the
prototype updating process the fact that the prototypes are assumed to be one
of the data objects is taken into consideration. Therefore, when the credal
partition matrix $\bm{M}$ is fixed, the new prototype of each cluster can be
obtained in a simpler manner than in the case of ECM application. The sECMdd algorithm is summarized as Algorithm \ref{alg:method}.
\begin{algorithm}\caption{\textbf{:} ~~~sECMdd  algorithm}\label{alg:method}
\begin{algorithmic}
\STATE {\textbf{Input:} Dissimilarity matrix $[\tau(x_i,x_j)]_{n\times n}$ for the $n$ objects
	$\{x_1,x_2,\cdots,x_n\}$.}
\STATE{\textbf{Parameters:}
	~\\$c$: number clusters $1<c<n$ \\ $\alpha$:
	weighing exponent for cardinality \\ $\beta >1$: weighting
	exponent \\ $\delta>0$: dissimilarity between any object to
	the empty set \\ $\eta>0$: to distinguish the outliers from the possible medoids\\
$\gamma \in [0,1]$: to balance of the contribution for imprecise classes\\}
\STATE {\textbf{Initialization:}\\ Choose
			   randomly $c$ initial prototypes from the object set
			}
\REPEAT
\STATE{
 (1). $t\leftarrow t+1$\\ (2). Compute $\bm{M}_t$ using
Eq.~\eqref{mass1}, Eq.~\eqref{mass2} and  $\bm{V}_{t-1}$\\ (3).
Compute the new prototype set $\bm{V}_{t}$ using Eq.~\eqref{pro_update} and \eqref{imepro}
}
\UNTIL{the prototypes remain unchanged.}
\STATE {\textbf{Output:} The optimal credal partition.}
\end{algorithmic}
\end{algorithm}

The update process of mass membership $\bm{M}$ is the same as that in ECM. For a given  $n \times n$ dissimilarity matrix,
the complexity of this step is of order $n 2^c$. The complexity for
updating the prototypes and calculating the dissimilarity between objects and classes is  $O(c n^2 + n  2^c)$.
Therefore, the total time complexity for one iteration in sECMdd is $O(cn^2+n 2^c)$.
%It is linear in the number of objects but exponential in the number of clusters.

\noindent \textbf{Remark 2:} The assignment update process will not increase $J_\text{sECMdd}$  since the new mass matrix is determined by differentiating of
the respective Lagrangian of the cost function with respect to $\bm{M}$. Also $J_\text{sECMdd}$ will not increase through the medoid-searching scheme for prototypes of specific classes. If the prototypes of specific classes are fixed, the medoids of imprecise classes determined by Eq.~\eqref{imepro} are likely to locate near to the ``centroid" of all the prototypes of the included specific classes. If the objects are in Euclidean space, the medoids of imprecise classes are near to the centroids found in ECM. Thus it  will not increase the value of the objective function also. Moreover,
the bba $\bm{M}$ is a function of the prototypes $\bm{V}$ and for given $\bm{V}$ the
assignment $\bm{M}$ is unique. Because sECMdd assumes that the prototypes are in the
original object data set $\bm{X}$, so there is a finite number of
different prototype vectors $\bm{V}$ and so is the number of corresponding credal
partitions $\bm{M}$. Consequently we can conclude that the sECMdd algorithm
converges in a finite number of steps.

\section{ECMdd with multiple weighted medoids}
This section presents evidential $c$-medoids algorithm using multiple weighted medoids. The approach to compute the relative weights of
 medoids is based on both the computation of the membership degree of objects belonging to
specific classes and the computation of the dissimilarities between objects.
\subsection{The objective function}
The objective function of wECMdd, $J_\text{wECMdd}$, has the same form as that in sECMdd (see Eq.~\eqref{costfun}). In wECMdd, we use multiple weighted medoids to represent each specific class instead of a single medoid. Thus the method to calculate $d_{ij}$ in the objective function is different from sECMdd. Let $\bm{V}^\Omega=\{v_{ki}^\Omega\}_{c\times n}$ be the weight matrix for specific classes, where $v_{ki}^\Omega$ describes the weight of object $i$ for the $k_{th}$ specific class. Then, the dissimilarity between object $x_i$ and cluster $A_j = \{\omega_k\}$ could be calculated by
\begin{equation}
  d(x_i, A_j)\triangleq d_{ij}=\sum_{l=1}^n \left(v_{kl}^\Omega\right)^\psi \tau(i,l),
\end{equation}
with
\begin{equation}
  \sum_{l=1}^n v_{kl}^\Omega = 1, \forall k = 1, 2, \cdots, c.
\end{equation}
Parameter $\psi$ controls the smoothness of the distribution of prototype weights.  The weights of imprecise class $A_j$ ($|A_j|>1$) can be derived according to the involved specific classes. If object $x_i$ has similar weights for specific classes $\omega_m$ and $\omega_n$, it is most probable  that $x_i$ lies in the overlapping area between two classes. Thus  the variance of the weights of object $x_i$ for all the included specific classes of $A_j$, $\mathrm{Var}_{ji}$, could be used to express the weights of $x_i$ for $A_j$ (denoted by $v^{2^\Omega}_{ji}$, and  $\bm{V}$ is used to denote the corresponding weight matrix\footnote{In sECMdd, $\bm{V}$ denotes the set of prototypes of all the classes. Here $\bm{V}$ represents the weights of prototypes. We use the same notation to show the  similar role of $\bm{V}$ in sECMdd and wECMdd. In fact sECMdd can be regarded as a special case of wECMdd, where the weight values are restricted to be either 0 or 1.}). The smaller $\mathrm{Var}_{ji}$ is, the higher $v^{2^\Omega}_{ji}$ is. However, we should pay attention to the outliers. They may hold similar small weights for each specific class, but have no contribution to the imprecise classes at all.  The minimum of $x_i$'s weights for all the associated specific classes could be taken into consideration to distinguish the outliers. If the minimal weight is too small, we should assign a small weight value for that object. Based on the discussion, the weights of object $x_i$ for class $A_j$ ($A_j \subseteq \Omega$) could be calculated as
\begin{equation}
  v^{2^\Omega}_{ji} = \frac{f_1\left(\mathrm{Var}\left(\{v^\Omega_{ki}; \omega_k \in A_j\}\right)\right) \cdot f_2\left(\mathrm{min}\left(\{v^\Omega_{ki}; \omega_k \in A_j\}\right)\right)}{\sum\limits_l f_1\left(\mathrm{Var}\left(\{v^\Omega_{kl}; \omega_k \in A_j\}\right)\right) \cdot f_2\left(\mathrm{min}\left(\{v^\Omega_{kl}; \omega_k \in A_j\}\right)\right)},
\end{equation}
where $f_1$ is a monotone decreasing function while $f_2$ is an increasing function. The two functions should be determined according to the application under concern. Based on our experiments, we suggest adopting the simple directly   and inversely proportion functions, {\em i.e.}
\begin{equation}
\label{vjiimprecise}
  v^{2^\Omega}_{ji} =  \frac{[\mathrm{min}\left(\{v^\Omega_{ki}; \omega_k \in A_j\}\right)]^{\xi}/\mathrm{Var}\left(\{v^\Omega_{ki}; \omega_k \in A_j\}\right)}{\sum\limits_l [\mathrm{min}\left(\{v^\Omega_{kl}; \omega_k \in A_j\}\right)]^{\xi}/\mathrm{Var}\left(\{v^\Omega_{kl}; \omega_k \in A_j\}\right)}.
\end{equation}
Parameter $\xi$ is used to balance the contribution of $f_1$ and $f_2$. It is remarkable that  when $A_j=\{\omega_k\}$, that is to say $|A_j|=1$, $v^{2^\Omega}_{ji}=v^\Omega_{ki}$. Therefore, the dissimilarity between object $x_i$ and cluster $A_j$ (including both specific and imprecise classes) could be given by
\begin{equation}
\label{dijemmdd}
  d_{ij}=
    \sum_{l=1}^n \left(v^{2^\Omega}_{jl}\right)^\psi \tau(i,l),  ~~~ A_j \subseteq \Omega, A_j \neq \emptyset.
\end{equation}
\subsection{Optimization}
The problem of finding optimal cluster assignments of objects and representatives of classes is now formulated as a constrained optimization problem, {\em i.e.} to find optimal values of $\bm{M}$ and $\bm{V}$ subject to a set of constrains. As before, the method of Lagrange multipliers could be utilized to derive the solutions. The Lagrangian function is constructed as
\begin{equation}
  L_\text{wECMdd}=J_\text{wECMdd}-\sum_{i=1}^n \lambda_i \left(\sum_{A_j\subseteq \Omega, A_j \neq \emptyset} m_{ij}-1 \right)-\sum_{k=1}^c \beta_k \left(\sum_{i=1}^n v^\Omega_{ki}-1\right),
\end{equation}
where $\lambda_i$ and $\beta_k$ are Lagrange multipliers. By calculating the first order partial derivatives of $L_\text{wECMdd}$ with respect to $m_{ij}$, $v^\Omega_{ki}$, $\lambda_i$ and $\beta_k$ and letting them to be 0, the update equations of $m_{ij}$ and $v^\Omega_{ki}$ could be derived. It is easy to obtain that the update equations for $m_{ij}$ are the same as Eqs.~\eqref{mass1} and \eqref{mass2} in the application of sECMdd, except that  in this case $d_{ij}$ should be calculated by Eq.~\eqref{dijemmdd}. The update strategy for the prototype weights $v^\Omega_{ki}$ is difficult to get since it is a non-linear optimization problem. Some specifical techniques may be adopted to solve this problem. Here we use a simple approximation scheme to update $v^\Omega_{ki}$.

Suppose the class assignment $\bm{M}$ is fixed and assume that the prototype weights for imprecise class $A_j$ ($A_j\subseteq \Omega, |A_j|>1$), $v^{2^\Omega}_{ji}$,  are dependent of the weights for specific classes ($v^\Omega_{ki}$). Then the first order necessary condition with respect to $v^\Omega_{ki}$ is only related to $d_{ij}$ with $A_j=\{\omega_k\}$. The update equations of $v^\Omega_{ki}$ could then derived as
\begin{equation}
\label{proweispe}
  v^\Omega_{ki}=\frac{\left(\sum\limits_{l=1}^n m_{lj}^\beta \tau_{li}\right)^{-1/(\psi-1)}}{\sum\limits_{h=1}^n\left(\sum\limits_{l=1}^n m_{lj}^\beta \tau_{lh}\right)^{-1/(\psi-1)}} ~~~ k=1,2,\cdots,c, ~A_j=\{\omega_k\}.
\end{equation}
After obtaining the weights for specific classes, the weights for imprecise classes can be obtained
 by Eq.~\eqref{vjiimprecise} and the dissimilarities
between objects and classes could then calculated by Eq.~\eqref{dijemmdd}. The  update of cluster assignment $\bm{M}$ and prototype weight matrix
$\bm{V}$ should be repeated until convergence. The wECMdd algorithm is summarised in Algorithm \ref{algo1}. The complexity of wECMdd is
$O(n 2^c + n^2)$.
% or reaching the maximum of iterations.

\begin{algorithm}\caption{\textbf{:} ~~~wECMdd  algorithm}\label{algo1}
\begin{algorithmic}
\STATE {\textbf{Input:} Dissimilarity matrix $[\tau(x_i,x_j)]_{n\times n}$ for the $n$ objects
	$\{x_1,x_2,\cdots,x_n\}$.}
\STATE{\textbf{Parameters:}
	~\\$c$: number clusters $1<c<n$ \\ $\alpha$:
	weighing exponent for cardinality \\ $\beta >1$: weighting
	exponent \\ $\delta>0$: dissimilarity between any object to
	the empty set \\
$\xi>0$: balancing the weights of imprecise classes\\
$\psi$: controlling the smoothness of the distribution of prototype weigths\\}
\STATE {\textbf{Initialization:}\\ Choose
			   randomly $c$ initial prototypes from the object set
			}
\REPEAT
\STATE{
(1). $t\leftarrow t+1$\\
(2). Compute $\bm{M}_t$ using
Eq.~\eqref{mass1}, Eq.~\eqref{mass2} and  $\bm{V}_{t-1}$\\
(3). Compute the prototype weights  for specific classes using Eq.~\eqref{proweispe}\\
(4). Compute the prototype weights for imprecise classes using Eq.~\eqref{vjiimprecise} and get the new $\bm{V}_{t}$. \\
}
\UNTIL{the prototypes remain unchanged.}
\STATE {\textbf{Output:} The optimal credal partition.}
\end{algorithmic}
\end{algorithm}

\noindent\textbf{Remark 3:} Existing work has studied the convergence properties of the partitioning clustering algorithms, such as $C$-Means, and $C$-Medoids. As we can see, wECMdd follows a similar clustering approach. The optimization process consists of three steps: cluster assignment update, prototype weights of specific classes update and then prototype weights of imprecise classes update.   The first two steps improve the objective function value by the application of Lagrangian multiplier method. The third step tries to find good representative objects for imprecise classes. If the method to determine the weights for imprecise classes is of practical meaning, it will also keep the objective function increasing. In fact the approach of updating the prototype weights  is similar to the idea of one-step Gaussian-Seidel iteration method, where the computation of the new variable vector uses the new elements that have already been computed, and the old elements  that have not yet to be advanced to the next iteration.  In Section \ref{expsec}, we will demonstrate through experiments that wECMdd could converge  in a few number of iterations.
\section{Application issues}
In this section, some problems when applying the ECMdd algorithms, such as how to adjust the parameters and how to select the initial prototypes  for each class, will be discussed.
\subsection{The parameters of the algorithm} As in ECM, before running ECMdd,
the values of the parameters have to be set. Parameters $\alpha, \beta$ and
$\delta$ have the same meanings as those in ECM. The value $\beta$ can be set to be $\beta=2$ in all experiments for which it
is a usual choice. The parameter $\alpha$  aims to penalize the subsets with
high cardinality and control  the amount of points assigned to imprecise
clusters for credal partitions.  The higher
$\alpha$ is, the less mass belief is assigned to the meta clusters and the
less imprecise will be the resulting partition. However, the decrease of
imprecision may result in high risk of errors. For instance, in the case of hard
partitions, the clustering results are completely precise but there is much more intendancy  to partition an object to
an unrelated group. As suggested in ~\citep{masson2008ecm}, a value can be used as a
starting default one but it can be modified according to what is expected from
the user.  The choice $\delta$ is more difficult and is strongly data
dependent~\citep{masson2008ecm}. If we do not aim at detecting outliers, $\delta$ can be set relatively large.

In sECMdd, parameter $\gamma$ weighs the contribution of  uncertainty to the dissimilarity between objects and imprecise clusters. %If the analyzed data set is highly overlapped, $\gamma$ could be set relatively small to make the imprecise classes express more uncertainty.
Parameter $\eta$ is used to distinguish the outliers from the possible medoids when determining the prototypes of meta classes. It can be set 1 by default and it has little effect on the final partition results. Parameters $\xi$ and $\psi$ are for specially for wECMdd. Similar to $\beta$, $\psi$ is used to control the smoothness of the weight
distribution. Parameter $\xi$ is used for not assigning  the outliers large weights for imprecise classes. If there are few outliers in the data set, it could be set to be near 0.

For determining the number of clusters, the validity index  of a credal
partition  defined by  \citet{masson2008ecm}
could be used:
\begin{equation}\label{clu_num_ind} N^*(c)\triangleq \frac{1}{n \log_2(c)}
	\times \sum_{i=1}^n \left[ \sum_{A\in 2^\Omega \setminus \emptyset}
	m_i(A)\log_2|A|+m_i(\emptyset)\log_2(c)\right], \end{equation}
where $0 \leq N^*(c) \leq 1$. This index has to be minimized to get the optimal number
of clusters.

As we discussed, in real practice, some of the parameters in the model such as $\beta, \eta$ and $\xi$ can be set as constants.  Although this
could not reduce the complexity of the algorithm,  it can  simplify the equations and bring about some convenience for applications.
\subsection{The initial prototypes}
The $c$-means type clustering algorithms are sensitive to the initial  prototypes \cite{celebi2013comparative}.
In this work, we follow the initialization procedure as the one used in \citep{gonzalez1985clustering, krishnapuram2001low, mei2010fuzzy} to generate a set of $c$ initial prototypes one by one. The first medoid, $\sigma_1$,  is randomly picked from the data set.
The rest of medoids are selected successively one by one in such a way that each one is most dissimilar to all the medoids that have already been picked. Suppose $\sigma=\{\sigma_1, \sigma_2, \cdots, \sigma_j\}$ is the  set of the first chosen $j$ ($j<c$) medoids. Then the $j+1$ medoid, $\sigma_{j+1}$, is set to the object $x_p$ with
\begin{equation}
 p = \arg\max\limits_{1\leq i \leq n; x_i \notin \sigma} \left\{\min\limits_{\sigma_k \in \sigma} \tau(x_i, \sigma_k)\right\}.
\end{equation}
This selection process makes the initial prototypes evenly distributed and locate as far away from each other as possible. It is noted that another scheme is that the first medoid is set to be  the object with the smallest total dissimilarity to all the other objects, {i.e.} $\sigma_1 = x_r$ with
\begin{equation}
   r= \arg \min\limits_{1 \leq i \leq n}\left\{\sum_{j=1}^n \tau(x_i, x_j)\right\},
\end{equation} and the remaining prototypes are selected the same way as before. \citet{krishnapuram2001low} have pointed out that both initialization schemes  work well in practice. But based on our experiments, for credal partitions, a bit of randomness of the first prototype might be desirable.

\subsection{Making the important objects more important}
In wECMdd, a matrix $\bm{V}=\{v^{2^\Omega}_{ji}\}$ %_{(2^c-1) \times n}$
 is used to
record prototype weights of $n$ objects with respect to all the  clusters, including the specific classes and imprecise classes. All objects are
engaged in describing clusters information  with some weights assigned to each detected classes. This seems unreasonable since it is easy to understand that when an object does not belong to a cluster, it should not participate in describing that cluster \citep{gao2014fuzzy}. Therefore, in each iteration of wECMdd, after the weights $v^\Omega_{ki}, k=1,2,\cdots,c, i=1,2,\cdots,n$ of $x_i$ for all the specific classes $\omega_k$ are obtained by Eq.~\eqref{proweispe}, the normalized weights $w^\Omega_{ki}$ could be calculated by \footnote{In the following we call this type of prototype weights ``normalized weights", and wECMdd with normalized weights is denoted by wECMdd-0. The standard wECMdd with multiple weights on all the objects described in the last section is still denoted by wECMdd.}
\begin{equation}
  w^\Omega_{ki} = \frac{v^{'}_{ki}}{\sum\limits_{i=1}^n v^{'}_{ki}}~, i=1,2,\cdots,n, ~\text{and}~ k=1,2,\cdots,c,
\end{equation}
where $v^{'}_{ki}$ equals to $v^\Omega_{ki}$ if $x_i$ belongs to $\omega_k$, 0 otherwise. Remark that $x_i$ is regarded as a member of class $\omega_k$ if $m_i(\{\omega_k\})$ is the maximum of the masses assigned to all the focal sets at this iteration.
%Then the weights of the objects for imprecise classes could be derived based on $w^\Omega_{ki}$ using Eq.~\eqref{vjiimprecise}. Overall, in this case, the weights of $x_i$ for all the focal sets are given as
%\begin{equation}
%  v^{2^\Omega}_{ji}=\begin{cases}
%    w^\Omega_{ki}, & \text{if}~~ A_j = \{\omega_k\}\\
%    \frac{[\mathrm{min}\left(\{w^\Omega_{ki}; \omega_k \in A_j\}\right)]^{\xi}/\mathrm{Var}\left(\{w^\Omega_{ki}; \omega_k \in A_j\}\right)}{\sum\limits_l [\mathrm{min}\left(\{w^\Omega_{kl}; \omega_k \in A_j\}\right)]^{\xi}/\mathrm{Var}\left(\{w^\Omega_{kl}; \omega_k \in A_j\}\right)}& \text{if} ~~|A_j|>1
%  \end{cases}.
%\end{equation}
%After that the dissimilarities between objects and clusters can be calculated consequently by Eq.~\eqref{dijemmdd}.
In fact, if we want to make the important ``core" objects more important in each cluster, a subset of fixed cardinality $1\leq q \ll n$ of objects $X$ could be used. The $q$ objects constitute core of each cluster, and collaborate to describe information of each class. This kind of wECMdd with $q$ medoids in each class is denoted by wECMdd-q. More generally, $q$ could be different for each cluster. However, how to determine $q$ or the number of cores in every class should be considered. This is not the topic of this work and we will study that in the future work.

\section{Experiments}
\label{expsec}
In this section some experiments on generated and real data sets will be performed to show the effectiveness of sECMdd and wECMdd. The results are compared with other relational clustering approaches PAM \citep{kaufman2009finding}, FCMdd \citep{krishnapuram2001low}, FMMdd \citep{mei2011fuzzy} and MECM \citep{zhou2015median} to illustrate the advantages of credal partitions and multi-prototype representativeness of classes. The  popular  measures, Precision (P), Recall (R) and Rand Index (RI),   which  are  typically  used  to  evaluate the
performance  of hard partitions are also used here. Precision  is the fraction  of relevant
instances (pairs in identical groups in  the clustering benchmark)  out  of
those  retrieved  instances (pairs in identical groups of the discovered
clusters), while recall is the fraction of relevant instances that are
retrieved. Then precision and recall can be calculated by
 \begin{equation}
	\label{precision} \text{P}=\frac{a}{a+c}
	~~~~~\text{and}~~~~~\text{R}=\frac{a}{a+d}
 \end{equation} respectively, where $a$ (respectively, $b$) be the number of pairs of
objects simultaneously assigned to identical classes (respectively, different
classes) by the stand reference partition and the obtained one. Similarly, values $c$ and $d$ are the numbers of dissimilar pairs  partitioned into the same cluster, and the number of similar object pairs clustered into
different clusters respectively.
The rand index  measures the percentage of correct decisions and it can be defined as
\begin{equation} \label{ri}
	\text{RI}=%\frac{\text{TP}+\text{TN}}{\text{TP}+\text{TN}+\text{FP}+\text{FN}}=
	\frac{2(a+b)}{n(n-1)},
 \end{equation}
where $n$ is the number of data objects.

For fuzzy and evidential clusterings,  objects  may  be partitioned into
multiple  clusters with different degrees.  In  such cases precision  would
be  consequently  low \cite{mendes2003evaluating}.
Usually the  fuzzy and
evidential clusters  are  made  crisp  before calculating  the  evaluation measures,
using  for  instance  the  maximum membership criterion
\cite{mendes2003evaluating} and pignistic probabilities
\cite{masson2008ecm}. Thus in this work we will
harden the fuzzy and credal clusters by maximizing the corresponding
membership and pignistic probabilities and calculate precision, recall and RI
for each case.

The introduced imprecise clusters can avoid the risk of partitioning a data into a
specific class without strong belief. In other words, a data pair can be
clustered into the same specific group only when we are quite confident and
thus the misclassification rate will be reduced. However, partitioning too
many data into  imprecise clusters may cause that  many objects are not
identified for their precise groups. In order to show the effectiveness of the
proposed method in these aspects, we use the indices for evaluating credal partitions, Evidential Precision (EP),
Evidential Recall (ER) and Evidential Rank Index (ERI) \cite{zhou2015median} defined as:
\begin{equation}\label{ep}
	\text{EP}=\frac{n_{er}}{N_e}, ~~~ \text{ER}=\frac{n_{er}}{N_r}, ~~~\text{ERI}=\frac{2(a^*+b^*)}{n(n-1)}.
\end{equation}
In Eq.~\eqref{ep}, the notation $N_e$ denotes the number of
pairs partitioned into the same specific group by evidential clustering, and
$n_{er}$ is the number of  relevant instance pairs out of these specifically
clustered pairs. The value $N_r$  denotes the number of pairs in the same
group of the clustering benchmark, and ER is the fraction of  specifically
retrieved instances (grouped into an identical specific cluster) out of these
relevant pairs. Value $a^*$ (respectively, $b^*$) is the number of pairs of
objects simultaneously clustered to the same specific class ({\em i.e.}
singleton class, respectively, different classes) by the stand reference
partition and the obtained credal one. When the partition degrades to a crisp one, EP, ER and ERI equal to
the classical precision, recall and rand index measures respectively.  EP and ER reflect
the accuracy of the credal partition from different points of view, but we
could not evaluate the clusterings from one single term. For example, if all
the objects are partitioned into imprecise clusters except two relevant data
object grouped into a specific class, $\text{EP}=1$ in this case. But we could
not say this is a good partition since it does not provide us with any information
of great value. At this time $\text{ER}\approx 0$. Thus ER could be used to express
the efficiency of the method for providing valuable partitions. ERI is like the combination of EP and ER
describing the accuracy of the clustering results.  Note that for  evidential  clusterings, precision, recall and RI  measures are
calculated after the corresponding hard partitions are obtained, while EP, ER and
ERI are based on hard credal partitions \cite{masson2008ecm}.

\subsection{Overlapped data sets}
\begin{center}\begin{figure}[!thbt] \centering
		\includegraphics[width=0.45\linewidth]{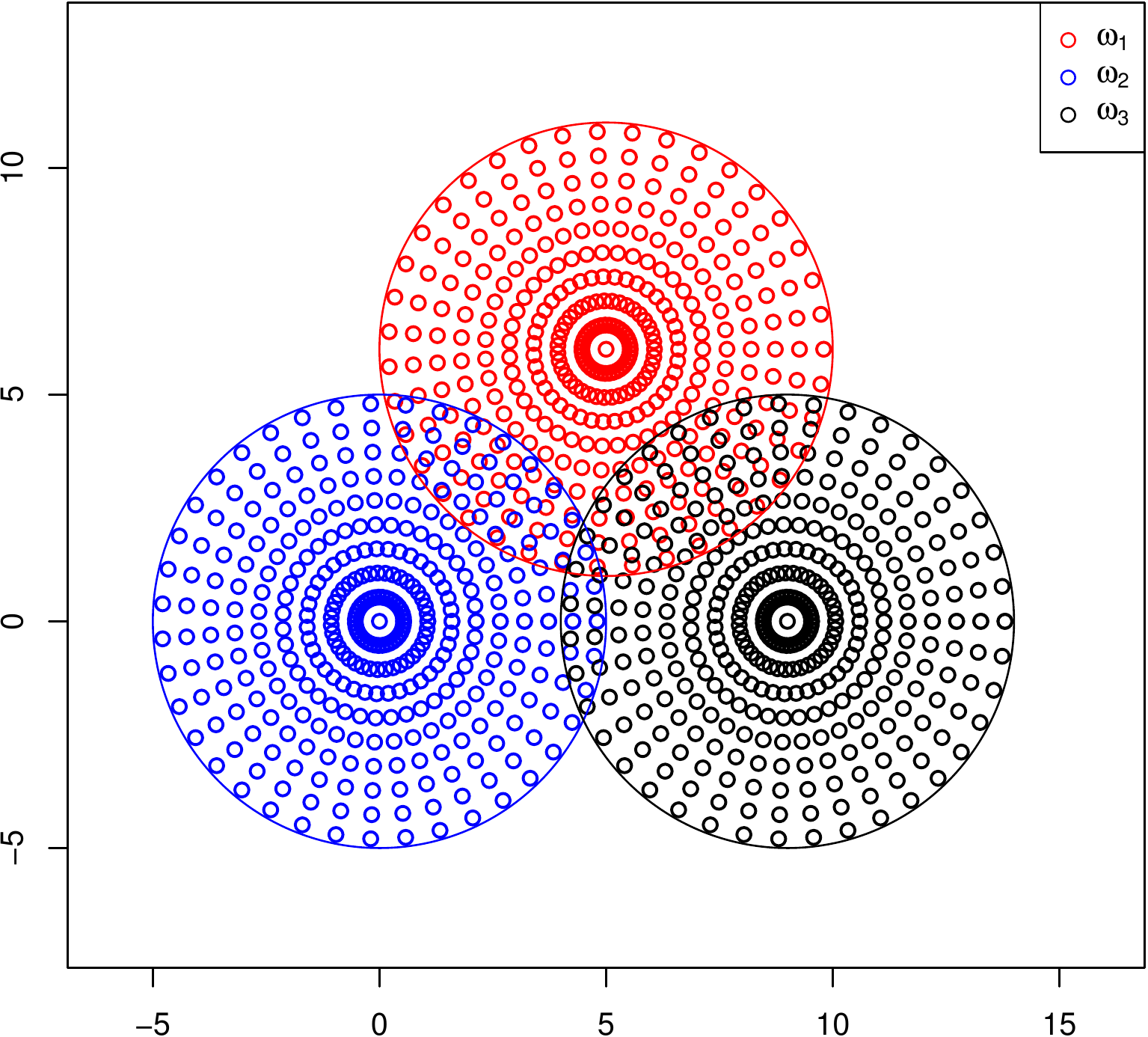}
		\hfill
		\includegraphics[width=0.45\linewidth]{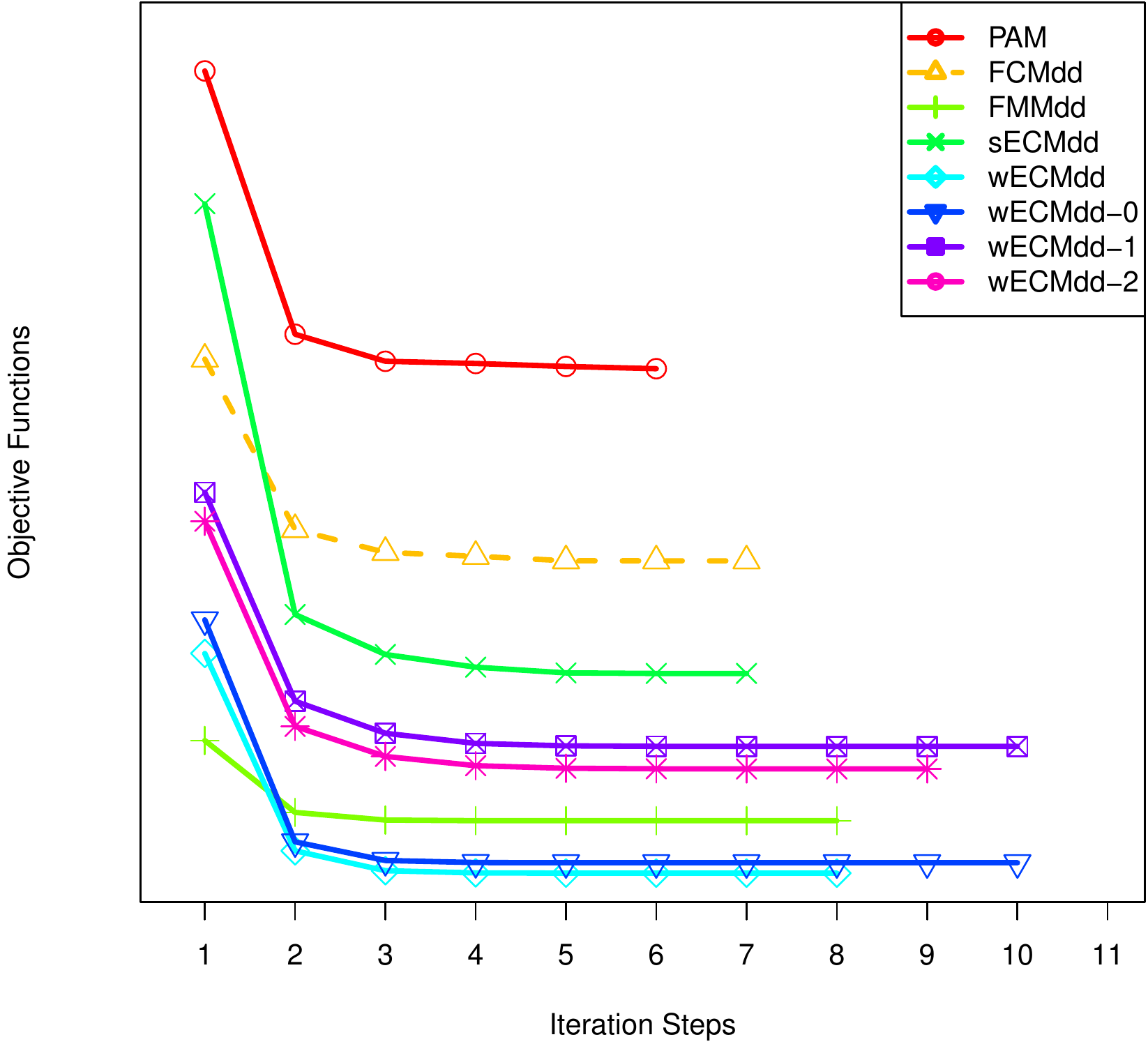}
		\hfill \parbox{.45\linewidth}{\centering\small a. Original data set} \hfill
		\parbox{.45\linewidth}{\centering\small b. Iteration steps}
		\hfill \caption{Clustering on overlapped data sets.}
	\label{rcircle_data}\end{figure} \end{center}

Due to the introduction of imprecise classes, credal partitions have the advantage to detect overlapped clusters. In the first example, we will
use  overlapped data sets to illustrate the behavior of the proposed algorithms. We start by generating  $3 \times 361$ points distributed in three overlapped circles with a same radius $R=5$ but with different  centers. The coordinates of
the first circle's center are $(5,6)$ while the coordinates of the other two circles' centers are $(0,0)$ and $(9,0)$. The data
set is displayed in Figure \ref{rcircle_data}.a.

Figure~\ref{rcircle_data}.b shows the iteration steps for different methods. For ECMdd clustering algorithms, there are three alternative steps to optimize the objective function (assignment update, and the update for medoids of specific and imprecise classes), while only two steps (update of membership and specific classes' prototypes) are required for the existing methods (PAM, FCMdd and FMMdd). But we can see from the figure, the added third step for calculating the new prototypes of imprecise classes in ECMdd clustering has no effect on the convergence.

\begin{center}\begin{figure}[!thbt] \centering
	\includegraphics[width=0.45\linewidth]{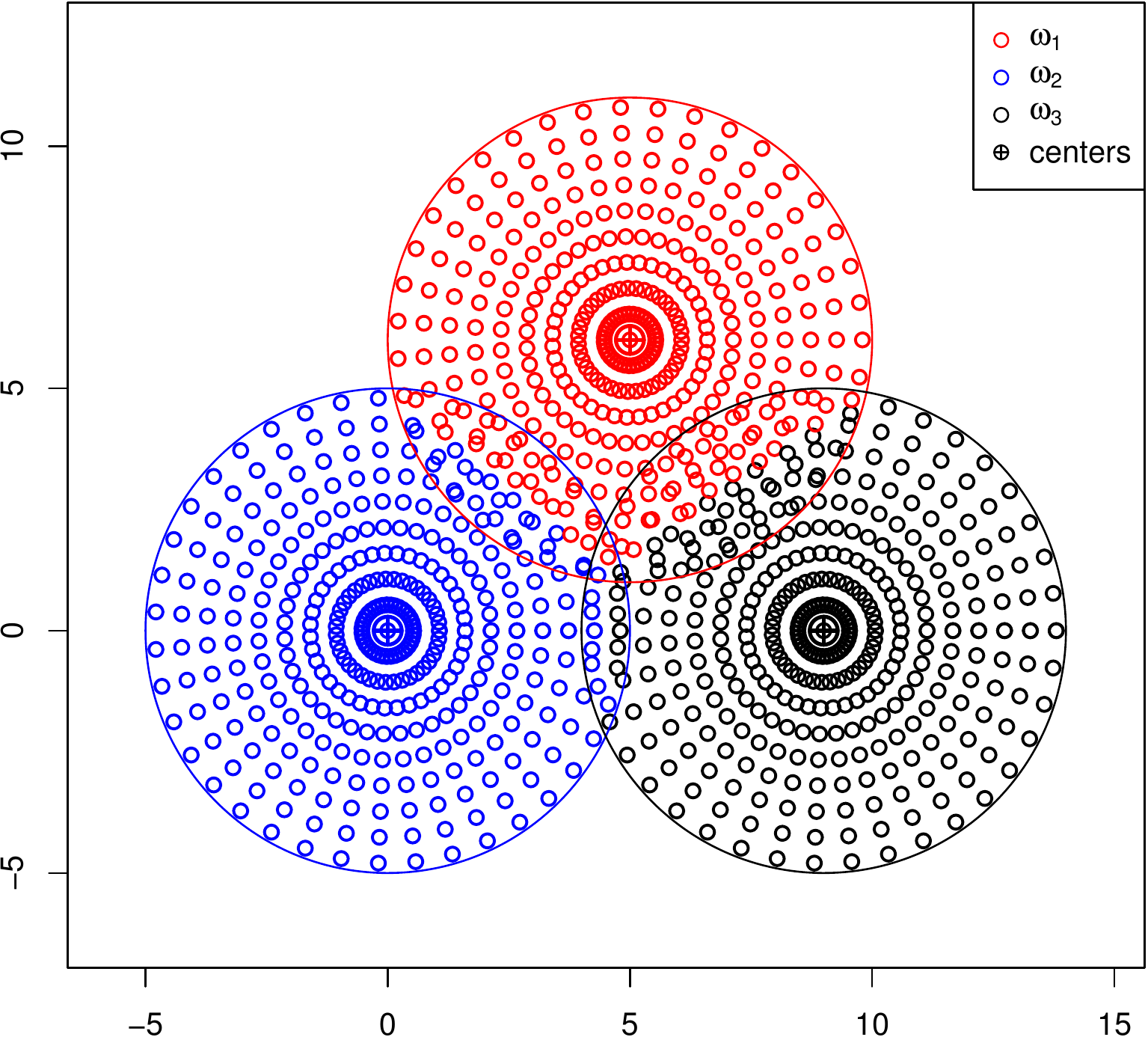}\hfill
	\includegraphics[width=0.45\linewidth]{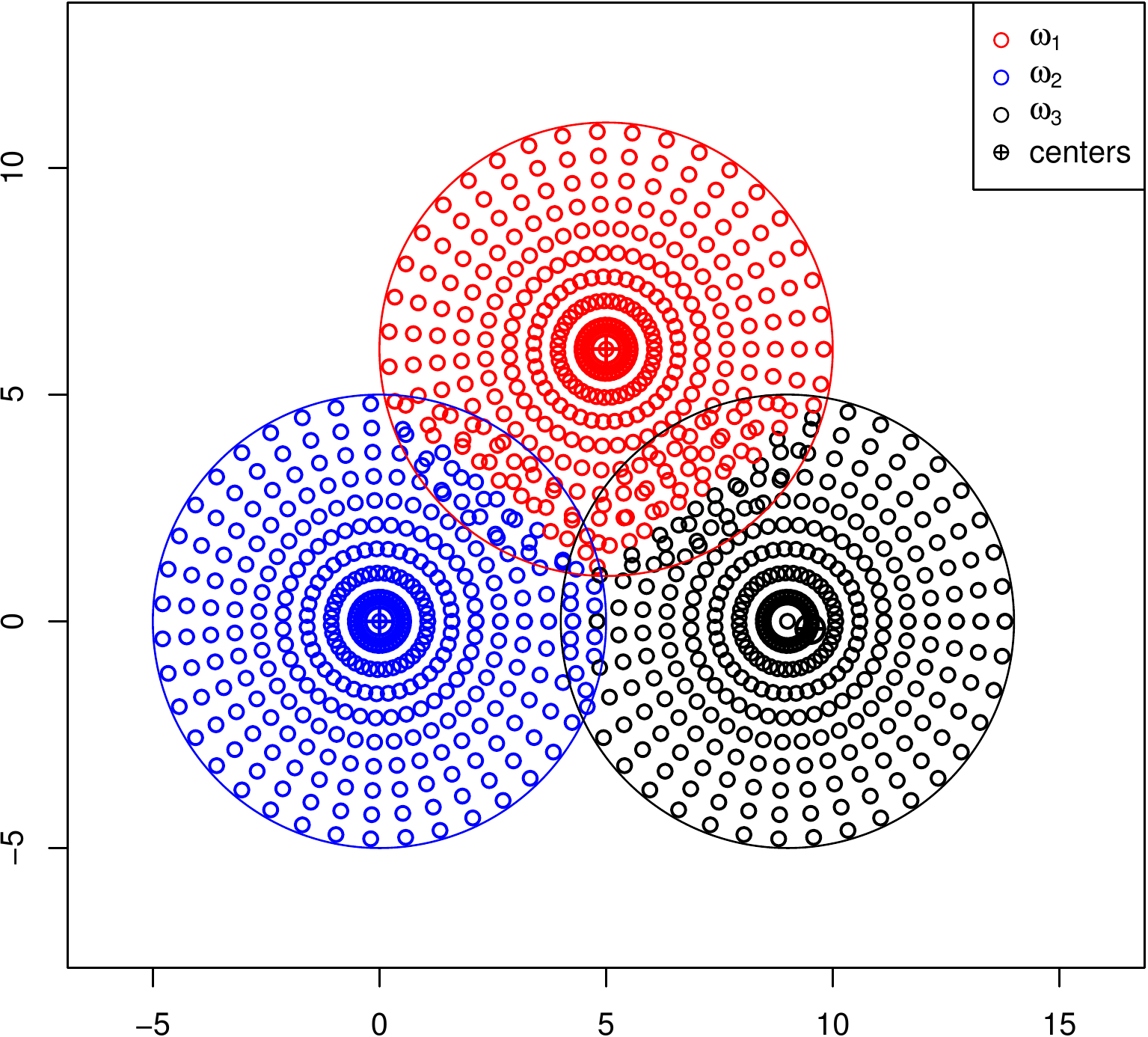}
		\hfill \parbox{.45\linewidth}{\centering\small a. PAM (FMMdd)} \hfill
		\parbox{.45\linewidth}{\centering\small b. FCMdd}
		%\includegraphics[width=0.45\linewidth]{circle3_fmmdd.pdf}\hfill
%	\includegraphics[width=0.45\linewidth]{circle3_secmdd.pdf}
%		\hfill \parbox{.45\linewidth}{\centering\small c. FMMdd} \hfill
%		\parbox{.45\linewidth}{\centering\small d. sECMdd}
	\includegraphics[width=0.45\linewidth]{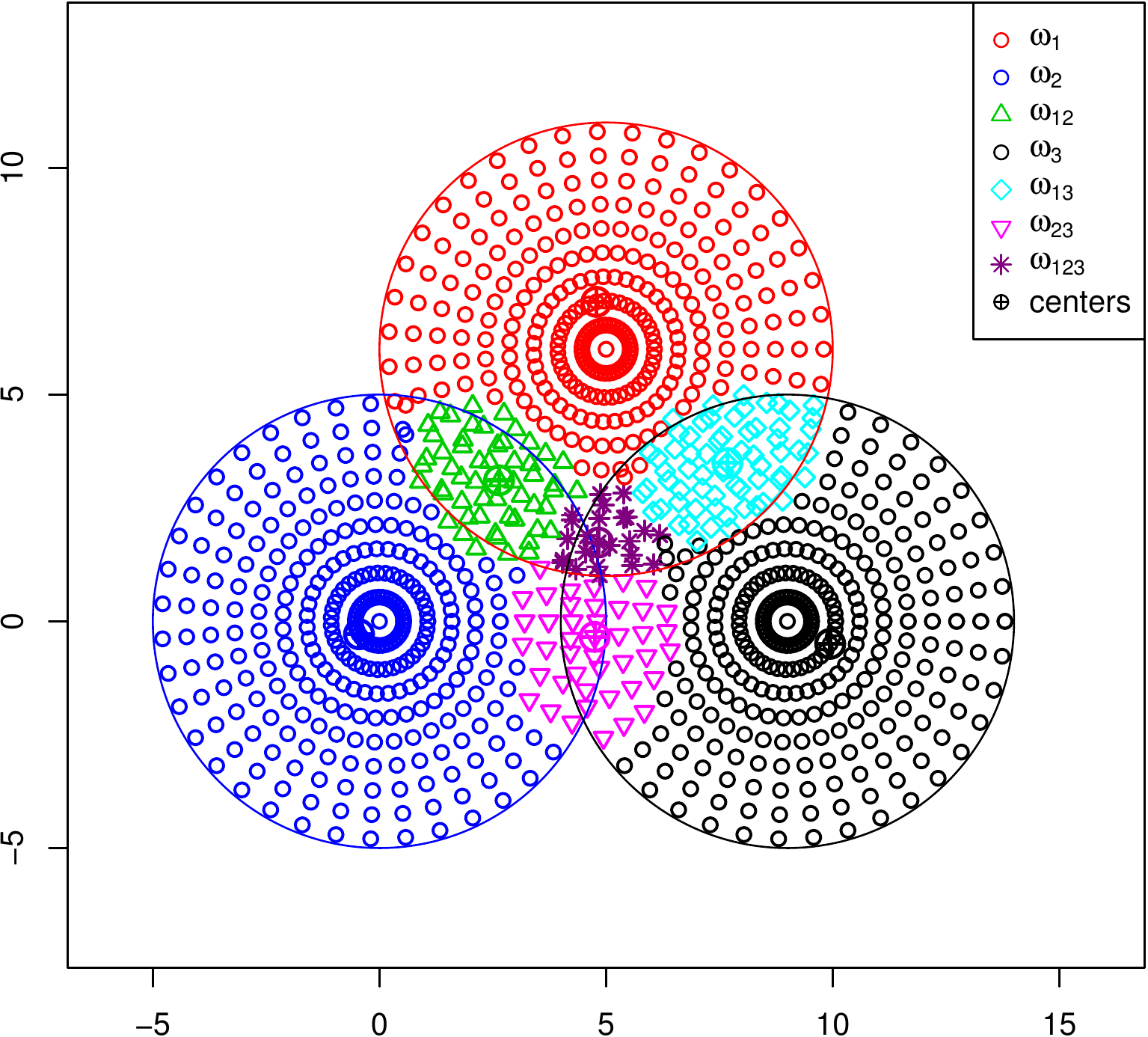}
		\hfill
		\includegraphics[width=0.45\linewidth]{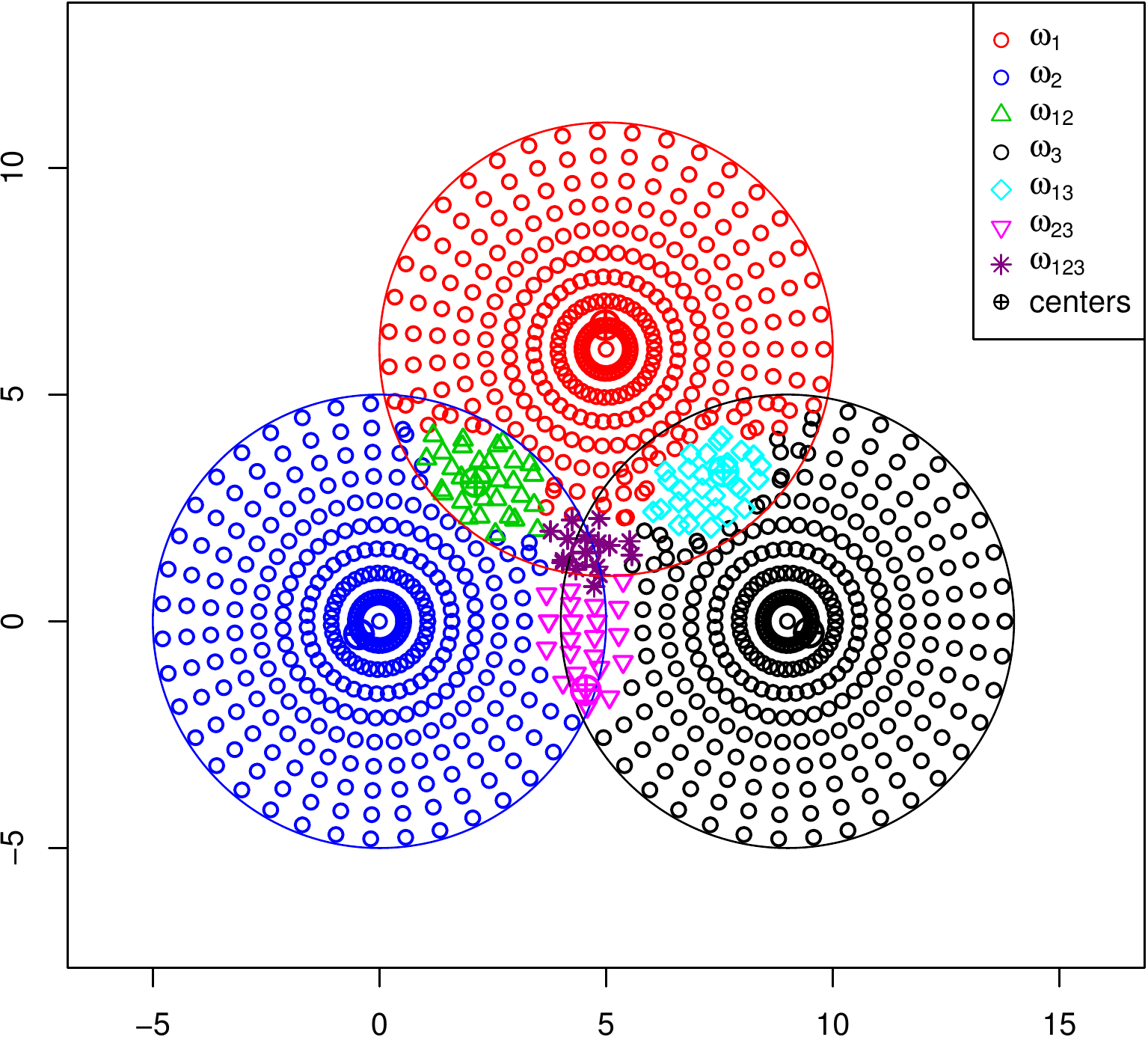}
		\hfill \parbox{.45\linewidth}{\centering\small c. sECMdd} \hfill
		\parbox{.45\linewidth}{\centering\small d. wECMdd (wECMdd-0)}
\includegraphics[width=0.45\linewidth]{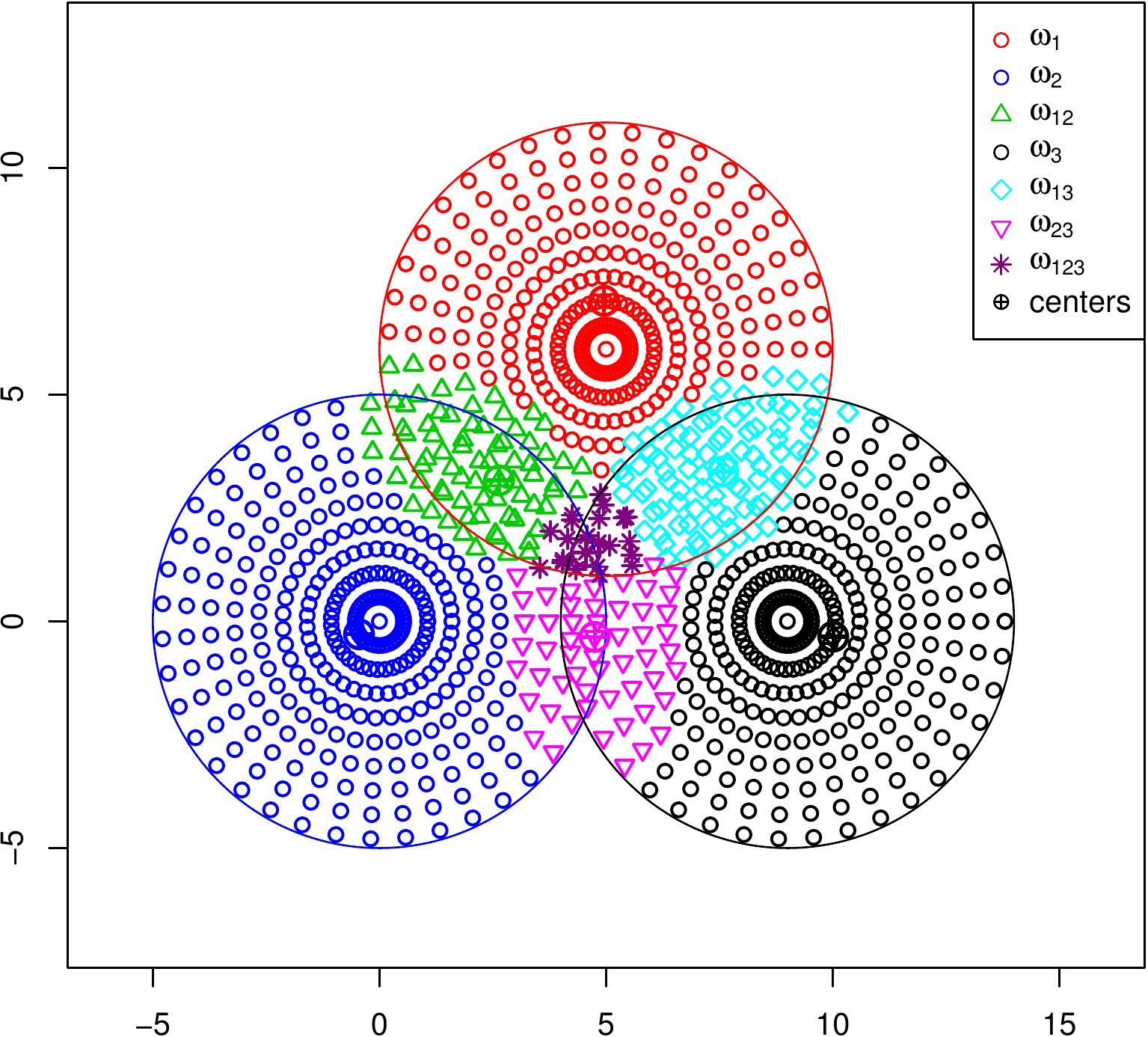}
		\hfill
		\includegraphics[width=0.45\linewidth]{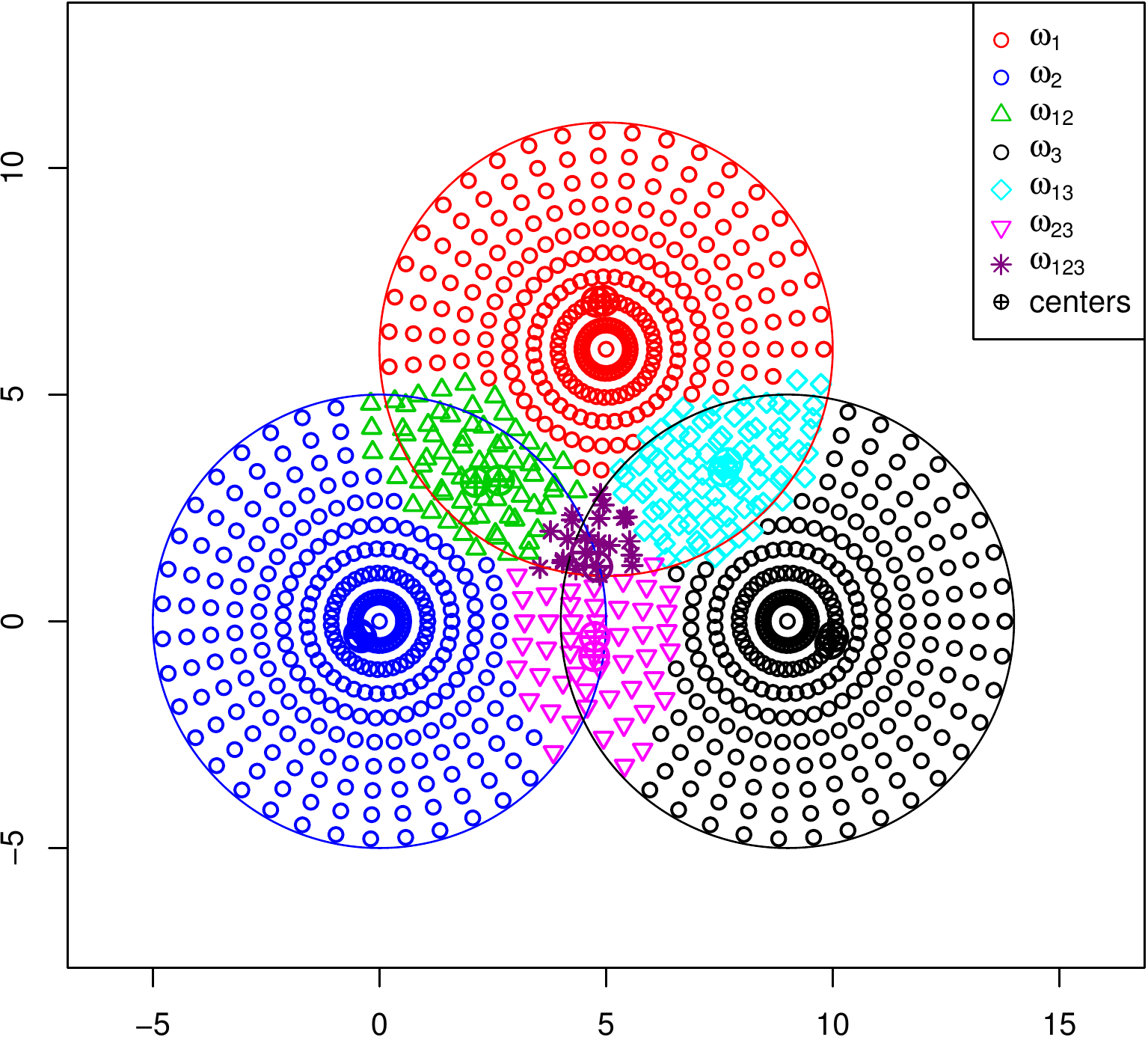}
		\hfill \parbox{.45\linewidth}{\centering\small e. wECMdd-1} \hfill
		\parbox{.45\linewidth}{\centering\small f. wECMdd-2}
		\hfill \caption{Clustering on overlapped data sets. All the methods are evoked with the same initial medoids. The prototypes in the detected classes by each method are marked with $\oplus$. For wECMdd and wECMdd-0,  the object with maximum weight in each class is marked as medoid. The results of PAM and FMMdd are similar, so we only display the figure of PAM to save space. And so also are the results for wECMdd and wECMdd-0.}
	\label{rcircle_data1}\end{figure} \end{center}

The fuzzy and credal partitions by different methods are shown in  Figure~\ref{rcircle_data1}, and the values of the evaluation indices are listed in Table \ref{circletable}. The objects are clustered into the class with the maximum membership values for fuzzy partitions (by FCMdd, FMMdd), while for credal partitions (by different ECMdd algorithms), with the maximum mass assignment. As a result, imprecise classes, such as $\{\omega_1, \omega_2\}$ (denoted by $\omega_{12}$ in the figure), are produced by ECMdd clustering to accept the objects for which it is difficult to make a precise (hard) decision. Consequently, the EP values of the credal partitions by ECMdd algorithms are distinctly high, which indicates that such soft decision mechanism could make the clustering result more ``cautious" and decrease the misclassification rate.

In this experiment, all the ECMdd algorithms are  run with: $\alpha=2, \beta = 2,  \delta = 100$. For sECMdd, $\eta=1$ and for wECMdd $\gamma = 1.2, \xi = 3$. The results by wECMdd and wECMdd-0 are similar, as they both use weights of objects to describe the cluster structure. The ECMdd algorithms using one (sECMdd, wECMdd-1) or two (wECMdd-2) objects to represent a class are sensitive to the detected prototypes. More objects that are not located in the overlapped area are inclined to be partitioned into the imprecise classes by these methods.
% latex table generated in R 3.1.1 by xtable 1.7-3 package
% Sun Mar 22 11:03:30 2015
\begin{table}[ht]
\centering \caption{The clustering results on the overlapped data set.}
\begin{tabular}{rlllllllllll}
  \hline
 & P & R & RI & EP & ER & ERI \\
  \hline
PAM & 0.8701 & 0.8701 & 0.9136 &0.8701 & 0.8701 & 0.9136\\
  FCMdd & 0.8731 & 0.8734 & 0.9156 & 0.8731 & 0.8734 & 0.9156 \\
  FMMdd & 0.8703 & 0.8702 & 0.9136 & 0.8703 & 0.8702 & 0.9136 \\
  sECMdd & 0.8715 & 0.8730 & 0.9149 & 0.9889 & 0.6799 & 0.8910 \\
  wECMdd & 0.8703 & 0.8705 & 0.9137 & 0.9726 & 0.7181 & 0.8994 \\
  wECMdd-0 & 0.8737 & 0.8738 & 0.9159 & 0.9405 & 0.7732 & 0.9083 \\
  wECMdd-1 & 0.8746 & 0.8764 & 0.9171 & 1.0000 & 0.6015 & 0.8674 \\
  wECMdd-2 & 0.8763 & 0.8780 & 0.9182 & 1.0000 & 0.6213 & 0.8740 \\
   \hline
\end{tabular}\label{circletable}
\end{table}

The running time of sECMdd, wECMdd, MECM, PAM, FCMdd, FMMdd is calculated to show the computational complexity\footnote{All the algorithms in this
work are implemented with R 3.2.1}.
Each algorithm is evoked 10 times with different initial parameters, and the average elapsed  time  is displayed in Table \ref{circletime}. As we can
see from the table,  ECMdd is of higher
complexity compared with fuzzy or hard medoid
based clustering. This is easy to understand,  as in the partitions there are imprecise classes and the membership is considered on the
extended frame of the power set $2^\Omega$. But credal partitions by the use of ECMdd will improve the precision
of the clustering results. This is also important in some applications, where  cautious
decisions are more welcome to avoid the possible high risk of misclassification.

% latex table generated in R 3.2.1 by xtable 1.7-4 package
% Fri Sep 25 10:50:12 2015
\begin{table}[ht]
\centering \caption{The average running time of different  algorithms.}
\begin{tabular}{rrrrrrr}
  \hline
 & sECMdd & wECMdd & MECM &PAM & FCMdd & FMMdd \\
  \hline
Elapsed Time (s) & 19.1100 & 14.2260 & 330.4680 &1.3000 & 1.3480 & 6.9080 \\
   \hline
\end{tabular}\label{circletime}
\end{table}

\begin{center}\begin{figure}[!thbt] \centering
		\includegraphics[width=0.45\linewidth]{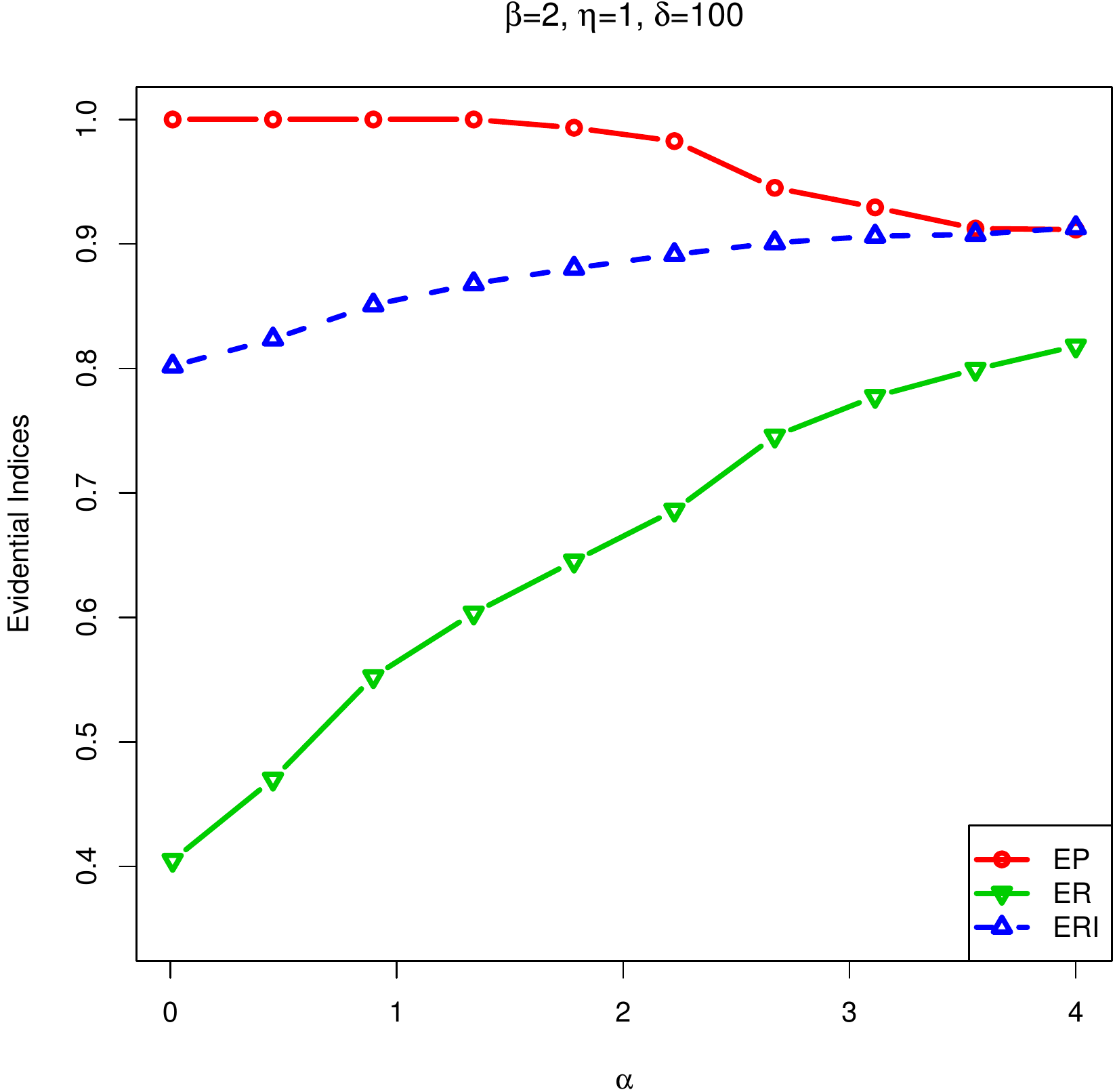}
		\hfill
		\includegraphics[width=0.45\linewidth]{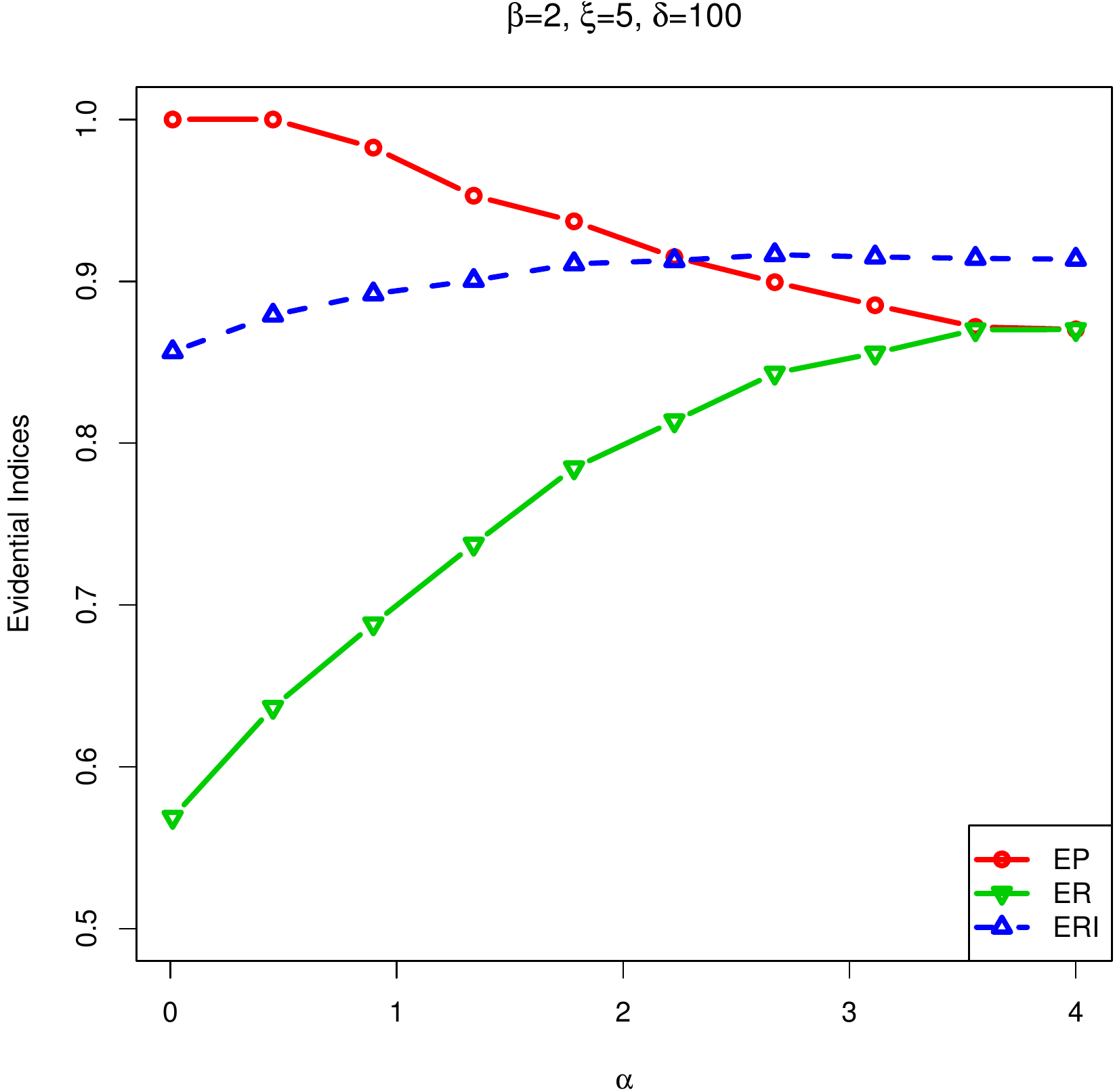}
		\hfill \parbox{.45\linewidth}{\centering\small a. sECMdd (with respect to $\alpha$)} \hfill
		\parbox{.45\linewidth}{\centering\small b. wECMdd (with respect to $\alpha$)}
	\includegraphics[width=0.45\linewidth]{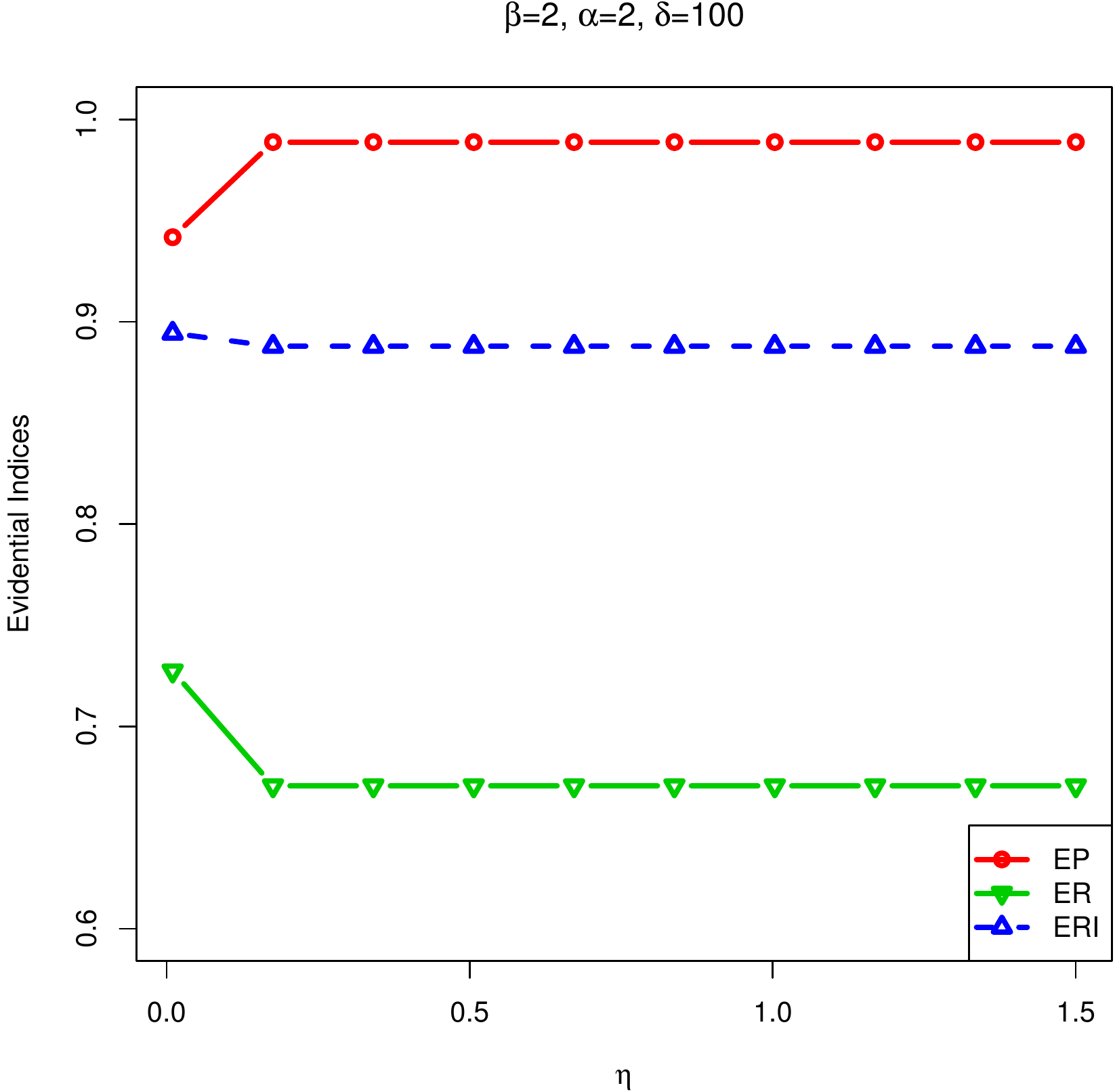}
		\hfill
		\includegraphics[width=0.45\linewidth]{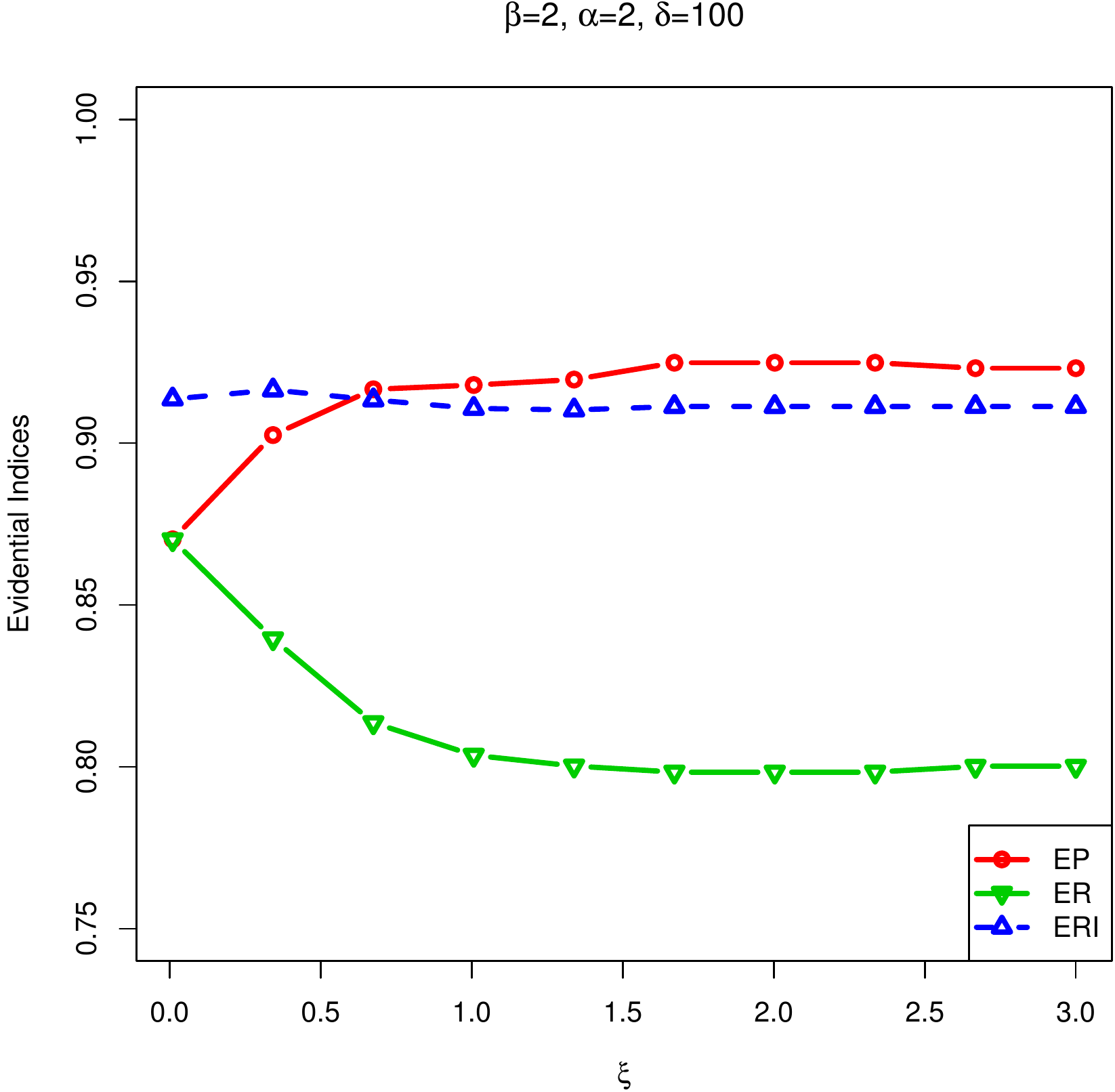}
		\hfill \parbox{.45\linewidth}{\centering\small c. sECMdd (with respect to $\eta$)} \hfill
		\parbox{.45\linewidth}{\centering\small d. wECMdd (with respect to $\xi$)}
	\includegraphics[width=0.45\linewidth]{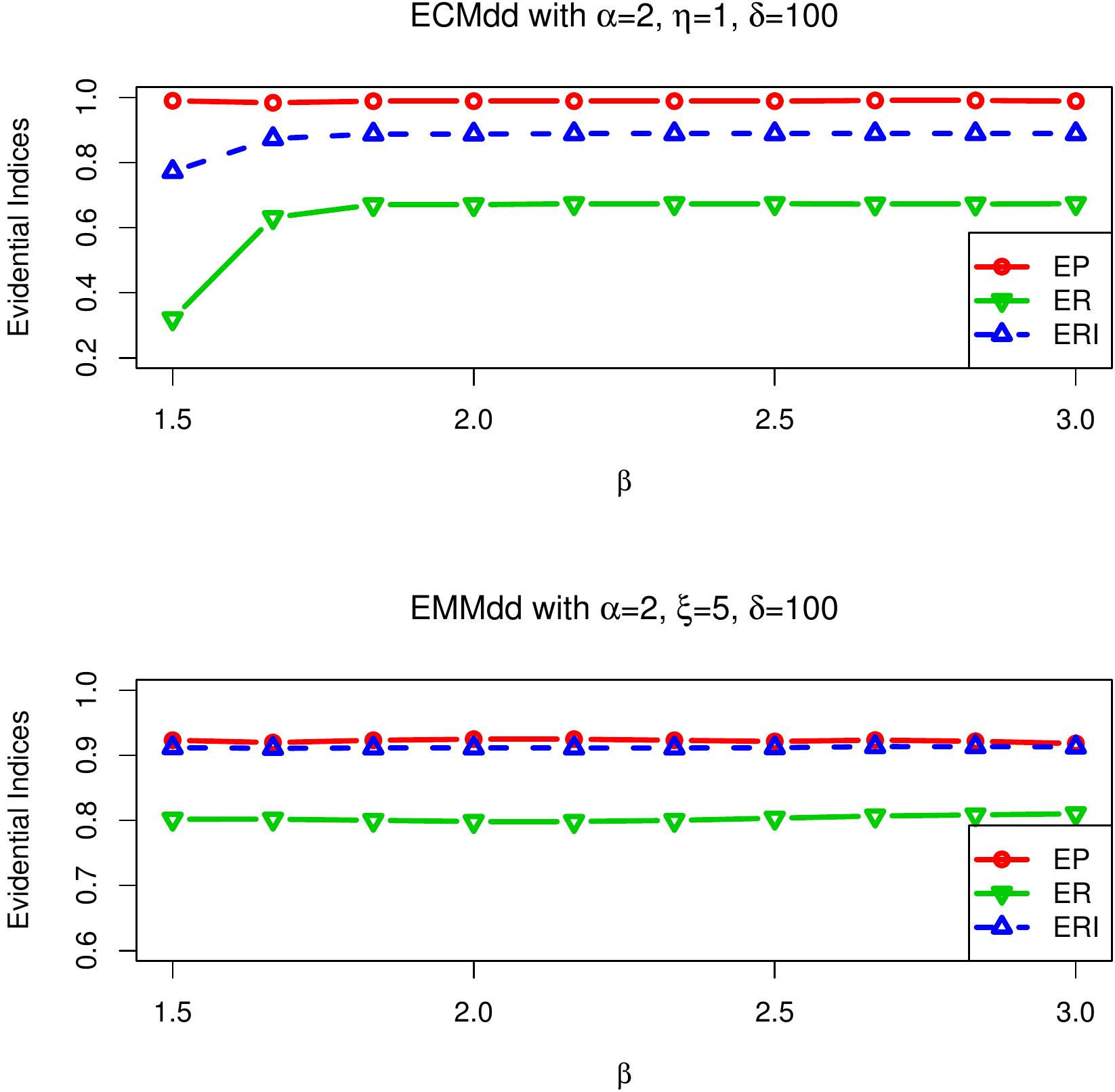}
		\hfill
		\includegraphics[width=0.45\linewidth]{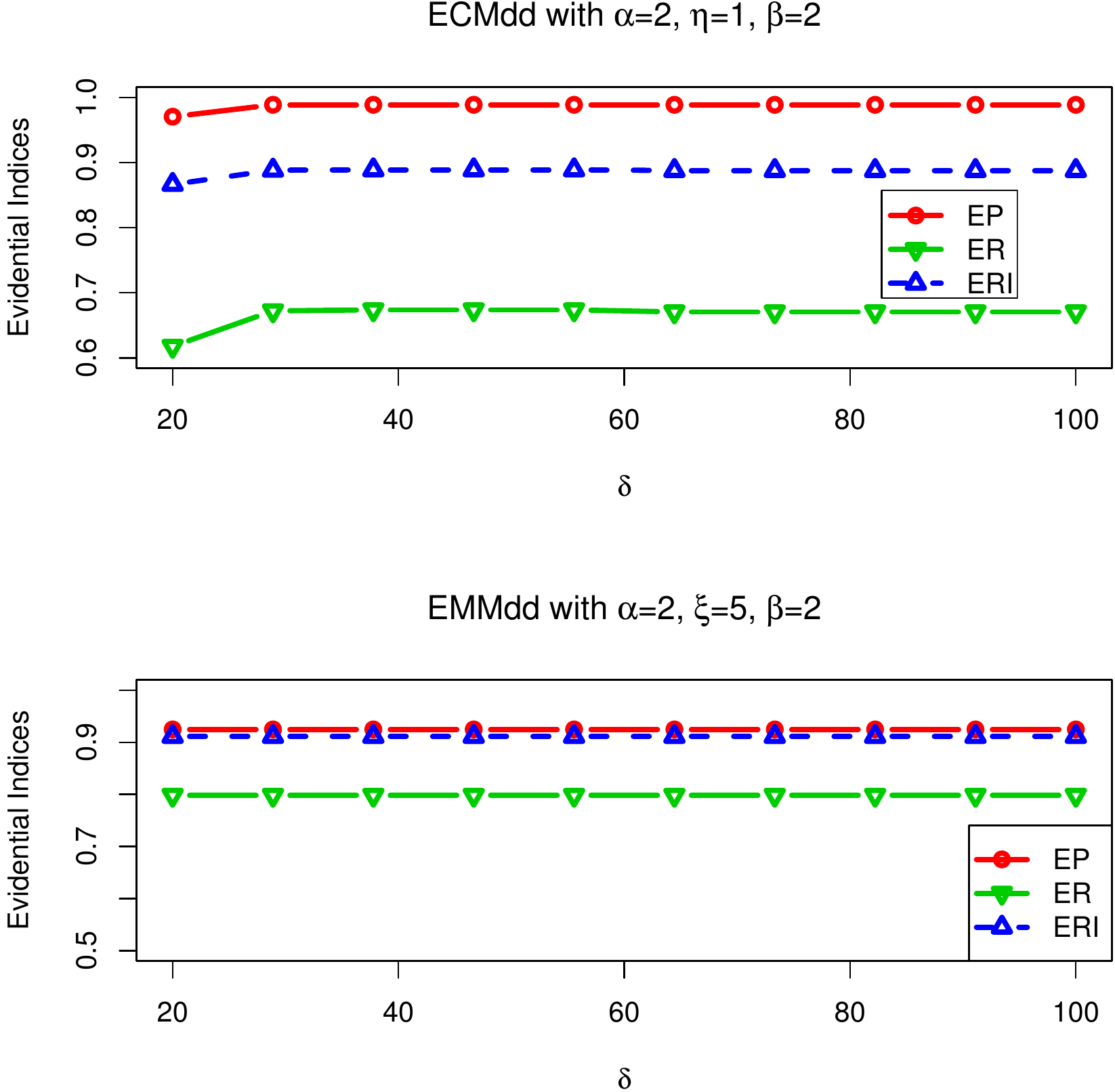}
		\hfill \parbox{.45\linewidth}{\centering\small e. sECMdd and wECMdd (with respect to $\beta$)} \hfill
		\parbox{.45\linewidth}{\centering\small f. sECMdd and wECMdd (with respect to $\delta$)}
		\hfill \caption{Clustering of overlapped data with different parameters.}
	\label{diffpara}\end{figure} \end{center}

In order to show the influence of parameters in ECMdd algorithms, different values
of $\alpha$, $\eta$, $\xi$, $\delta$ and $\beta$ have been tested for this data set.
Figure~\ref{diffpara}.a displays the three evidential indices varying with
$\alpha$ by sECMdd, while Figure~\ref{diffpara}.b
depicts the results of wECMdd with different $\alpha$. As we can see,
for both sECMdd and wECMdd, if we want to make more imprecise decisions to improve
ER, parameter $\alpha$ can be decreased, since $\alpha$ tries to adjust the penalty degree to control
the imprecise rates of the results. Keeping more soft decisions will reduce the misclassification rate and makes  the
specific decisions more accurate. But the partition results with few specific decisions have low ER values and they are of limited practical meaning. In application we should determine $\alpha$ based on the requirement. Parameter $\eta$ in sECMdd and $\xi$ in wECMdd are both for distinguish the outliers in imprecise classes.  As pointed out in
Figures \ref{diffpara}.c and  \ref{diffpara}.d, if $\eta$ and $\xi$ are well set, they have little
effect on the final clusterings. The same is true in the case of $\delta$ which is applied to detect outliers (see Figure \ref{diffpara}.f). The effect of various values of $\beta$ is
illustrated in Figure \ref{diffpara}-e. We can see that it has little influence on the
final results as long as it is larger than 1.7.  Similar to FCM and ECM,  the value of $\beta$   could also be set to be 2 as a  usual
choice here.

Although  there are a lot of  parameters to adjust in the proposed methods, but compared
with MECM (the discussion about the parameters of MECM could be seen in Ref.~\citep{zhou2015median}), the parameters of ECMdd
are much easier to adjust and control. In fact from the experiments we can see that only parameter $\alpha$ has a great influence
on the result. The other parameters such as $\beta$, $\eta$ (for sECMdd), $\xi$ (for wECMdd)
can be set as default  for simplicity. These parameters are involved in
the model in order to enhance the flexibility.
When the analyzed data set has high overlap, the value of $\alpha$ can be set
small to get more imprecise and cautious decisions with relatively high EP value.  However,
the improvement of precision will bring about the decline of
recall, as more data could not be clustered into specific classes.
What we should do is to set parameters based on our own requirement to make a tradeoff between precision and recall.  Values of these parameters can be also
learned from historical data if such data are available.

\subsection{Gaussian data set}
\begin{center}\begin{figure}[!thbt] \centering
 		\includegraphics[width=0.45\linewidth]{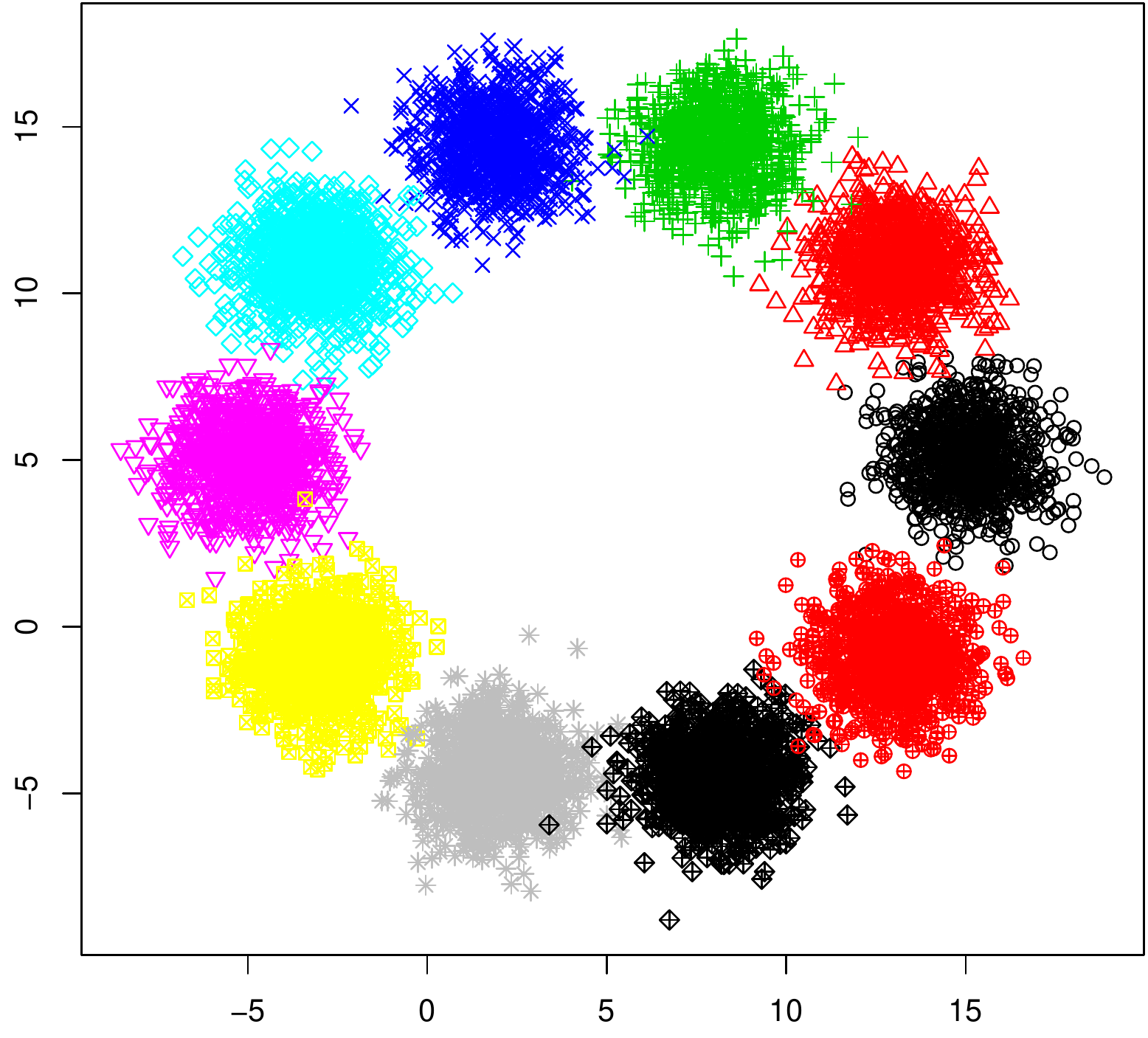}
		\hfill \caption{Gaussian data set.}
	\label{gaussdata}\end{figure} \end{center}

In the second experiment, we test on a data set consisting of 10000 points generated from different Gaussian distributions. The
points are from 10 Gaussian distributions, the mean values of which are uniformly located in a circle.  The data set is displayed
in Figure \ref{gaussdata}.

\begin{table}[ht]
\centering\caption{The clustering results on Gaussian data set.}
\begin{tabular}{rrrrrrrrrrrrrrrrrrrr}
  \hline
 & P & R & RI & EP & ER & ERI & Elapsed Time (s)\\
  \hline
PAM & 0.8939 & 0.8940 & 0.8988 & 0.8939 & 0.8940 & 0.8988 & 118.2097 \\
  FCMdd & 0.8960 & 0.8960 & 0.8992 & 0.8960 & \textbf{0.8960} & 0.8992 & 152.4320\\
  FMMdd &0.8928 & 0.8980 & 0.8996 & 0.8980 & 0.8928 & 0.8996 & 197.5340\\
  MECM  & \textbf{0.8980} & 0.8940 & 0.8921 & 0.9932 & 0.3173 & 0.9321 &19430.1560 \\
  sECMdd & 0.8931 & \textbf{0.8992} & \textbf{0.9043} & \textbf{1.0000} & 0.4468 & 0.9452  & 8987.7390\\
  wECMdd & 0.8923 & 0.8914 & 0.8908 & \textbf{1.0000} & 0.5623 & \textbf{0.9566} &8534.8740 \\
   \hline
\end{tabular}\label{gauss10000}
\end{table}

Table \ref{gauss10000} lists the indices for evaluating the
different methods. Bold entries in each column of this table (and also other tables in the following) indicate
that the results are significant as the top performing
algorithm(s) in terms of the corresponding evaluation index. We can see that the precision, recall and RI
values for all approaches are similar.  As the data objects are from gaussian
distributions, it is intuitive that there is only one geometrical center in each class. That's why the one-prototype
based clustering  sECMdd is a little better than wECMdd. For evidential clusterings, {\em e.g.,} MECM, sECMdd and wECMdd, the three
classical measures are based on the associated pignistic
probabilities.  It indicates that credal partitions can
provide the same information as crisp and fuzzy ones (PAM, FCMdd, and FMMdd). Most of the misclassifications in this
experiment come from the data points lying in the overlapped area between two classes.

However, from the same table, we
can also see that the evidential measures EP and ERI by sECMdd and wECMdd are higher (for hard
partitions, the values of evidential measures equal  to the  corresponding
classical ones) than  the ones obtained by other methods. This fact confirms the accuracy of the specific decisions
{\em i.e.} decisions clustering the objects into specific classes. The advantage can be attributed to
the introduction of imprecise clusters, with which we do not have to partition the
uncertain or unknown objects lying in the overlap into a specific cluster. Consequently, it could reduce the risk of
misclassification. For the computational time, the same conclusion  as in the first experiment can be obtained. Evidential clustering algorithms
(sECMdd, wECMdd and MECM) are more time-consuming than hard or
fuzzy ones. But we can see that wECMdd is the fastest one among the three, and it is significantly better than
MECM in terms of complexity.

\subsection{$X_{12}$ data set} 
\begin{center}\begin{figure}[!thbt] \centering
		\includegraphics[width=0.45\linewidth]{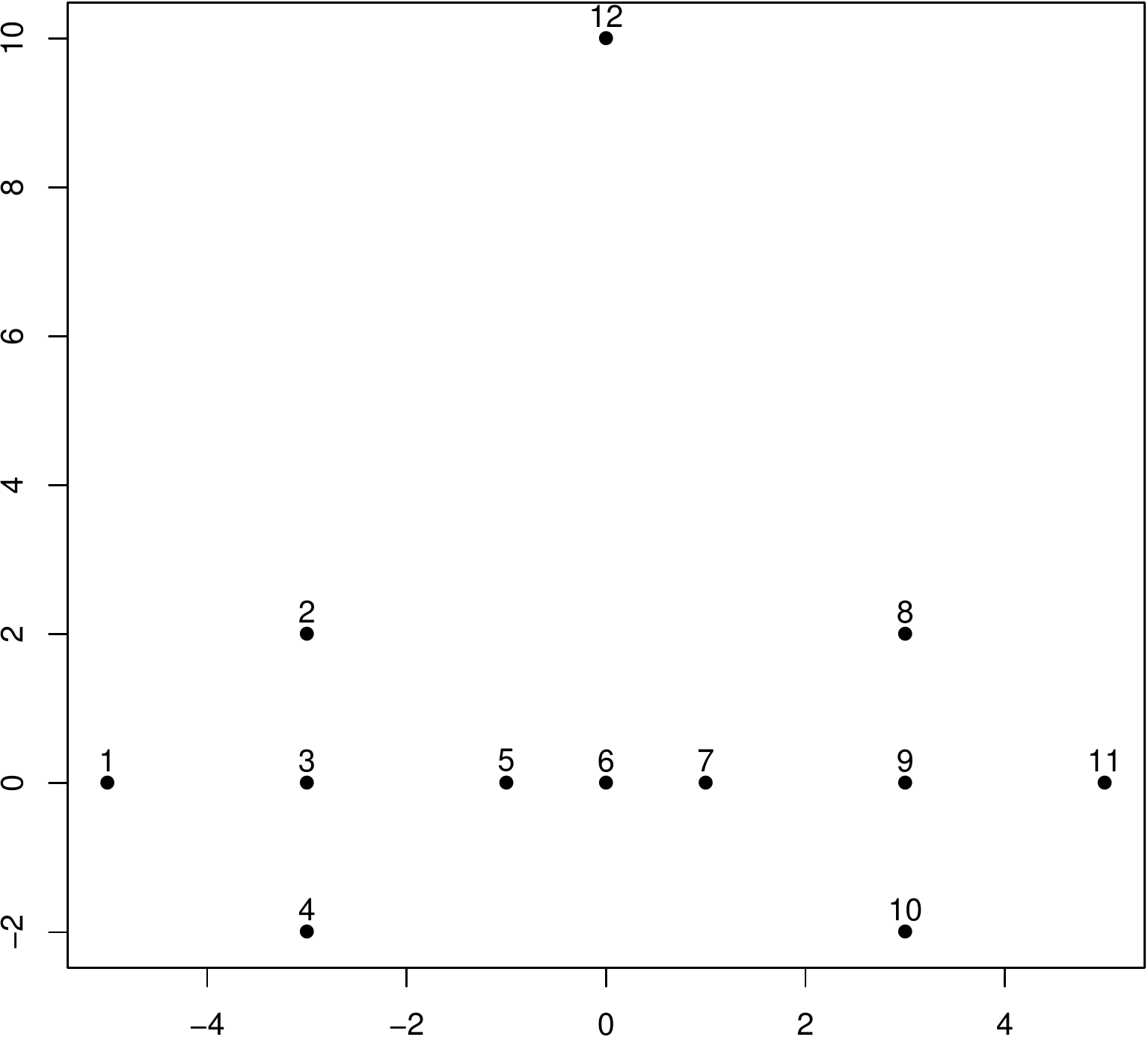}
		\hfill
		\includegraphics[width=0.45\linewidth]{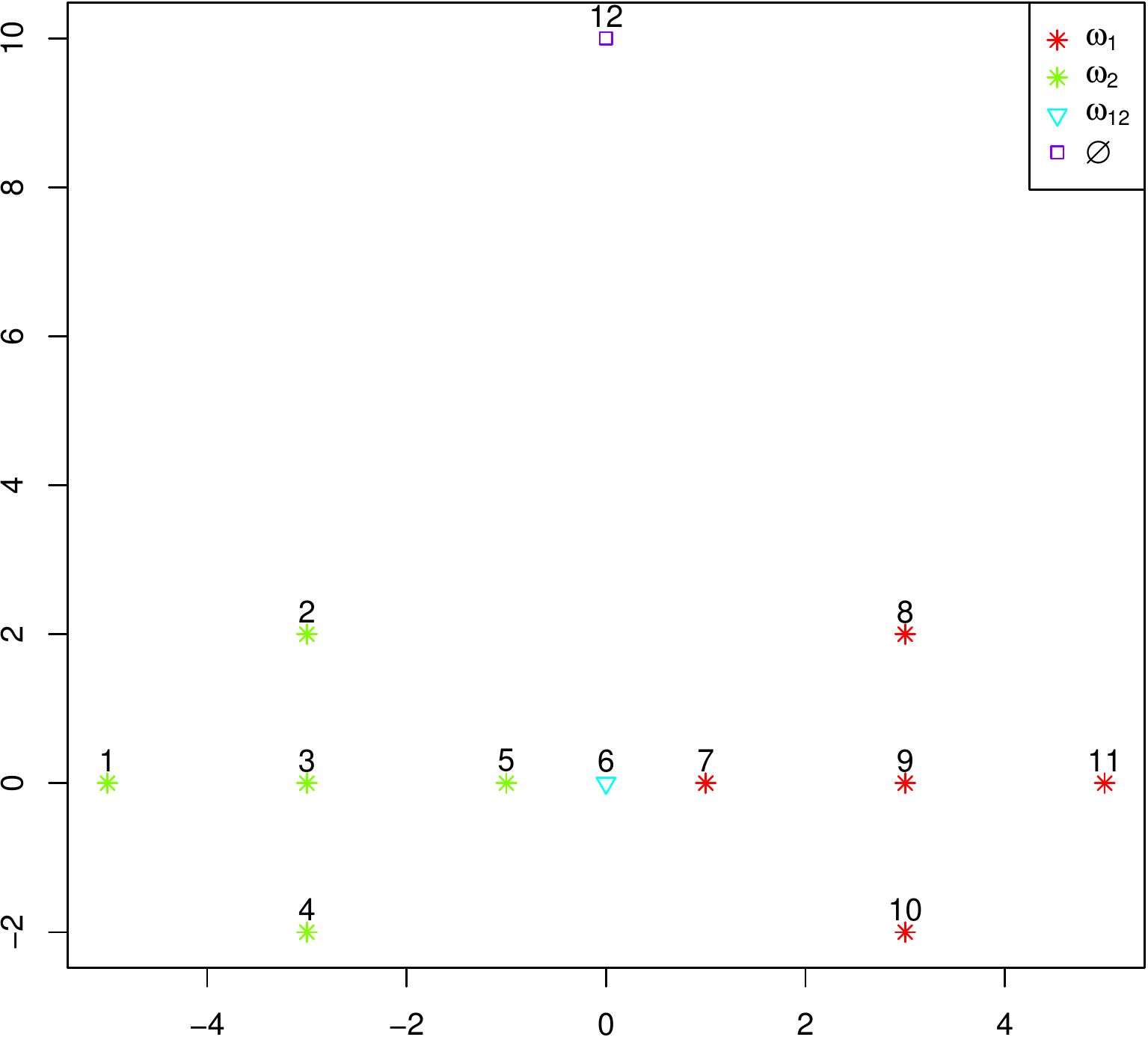}
		\hfill \parbox{.45\linewidth}{\centering\small a. Original data set} \hfill
		\parbox{.45\linewidth}{\centering\small b. sECMdd \& wECMdd}
		\hfill \caption{A simple data set of 12 objects.}
	\label{x12data}\end{figure} \end{center}

In this test,  a simple classical data set composed of 12 objects represented in Figure \ref{x12data}.a is considered. As we can see from the figure, objects 1 - 11 are clearly dived into two groups whereas object 12 is an outlier. The results by sECMdd and wECMdd are shown in Figure~\ref{x12data}.b.  Object 6 is clustered into imprecise class $\omega_{12}\triangleq\{\omega_1, \omega_2\}$ while object 12 is regarded as an outlier (belonging to  $\emptyset$).

In this data set, object 6 is a ``good" member for both classes,
whereas object 12 is a ``poor" point.  It can be seen from
Table~\ref{x12table} that the  fuzzy partition  by FCMdd also gives
large equal membership values to $\omega_1$ and $\omega_2$ for object 12, just
like in the case of such good members as point 6. The same is true for PAM and FMMdd. The obtained results show  the problem  of
distinguishing between ignorance and the ``equal evidence" (uncertainty) for fuzzy
partitions. But the table shows that the credal
partition by wECMdd assigns largest mass belief to $\emptyset$ for object 12, indicating it is an outlier. Moreover, the values $v^{2^\Omega}_{ji}$ in the table are the weights of object $i$ for class $A_j$, from which it can be seen that object 3 and object 9 play a center role in their own classes, while object 6 contributes most to the overlapped parts of the two classes. Thus the prototype weights indeed could provide us some rich information about  the cluster structure.

%ECMdd and wECMdd are run with the common following parameters: $\alpha=2,\beta=2,\delta=10$, while for wECMdd $\xi_1=2, \xi_2 =5$ and for ECMdd $\eta = 0.7$.
% latex table generated in R 3.1.1 by xtable 1.7-3 package
% Sun Feb 08 19:18:43 2015
% latex table generated in R 3.1.1 by xtable 1.7-3 package
% Sun Feb 08 19:19:56 2015
\vspace{-3em}
\begin{table}[ht]
\centering \caption{The clustering results of $X_{12}$  data set using FCMdd and wECMdd. The objects marked  with * are the medoids found by FCMdd.
Values $m_{ij}, j=1,2,3,4$ are the mass assigned to $x_i$ for class $\emptyset$, $\omega_1$, $\omega_2$ and imprecise class $\omega_{12}\triangleq \{\omega_1,\omega_2\}$. Values $v^{2^\Omega}_{ij}, j=1,2,3$ are the weights of object $x_i$ for class $\omega_1$, $\omega_2$ and $\omega_{12}$.}
\resizebox{\textwidth}{!}{
\begin{tabular}{l|ll|llll|ll|llllllllll}
\hline
& \multicolumn{2}{c|}{FCMdd} &\multicolumn{8}{c}{wECMdd}\\
  %\cline{2-4} \cline{6-14}
 \hline
 id & $u_{i1}$ &$u_{i2}$ & $m_{i1}$ & $m_{i2}$ & $m_{i3}$ & $m_{i4}$&$\BetP_{i1}$&$\BetP_{i2}$ & $v^{2^\Omega}_{1i}$ & $v^{2^\Omega}_{2i}$ & $v^{2^\Omega}_{3i}$ \\
  \hline
1 & 0.9412&0.0588 &   0.1054 & 0.7242 & 0.1599 & 0.0105 &0.8154&0.1846 & 0.1123 & 0.0230 & 0.0000 \\
  2 & 0.9091&0.0909 &   0.0749 & 0.7282 & 0.1825 & 0.0144 &0.7950&0.2050 & 0.1396 & 0.0359 & 0.0000 \\
  3 & 1.0000&0.0000*  &0.0502 & 0.8005 & 0.1354 & 0.0140 & 0.8501&0.1499 & \textbf{0.1829} & 0.0382 & 0.0000 \\
  4 & 0.9091&0.0909 &   0.0821 & 0.7083 & 0.1938 & 0.0158 &0.7803&0.2197 & 0.1117 & 0.0337 & 0.0000 \\
  5 & 0.8000&0.2000 &   0.0438 & 0.5969 & 0.2498 & 0.1095 &0.6815&0.3185 & 0.1386 & 0.0709 & 0.0001 \\
  6 & 0.5000&0.5000 &   0.0000 & 0.0000 & 0.0000 & 1.0000 &0.5000&0.5000 & 0.0997 & 0.0999 & \textbf{0.9998} \\
  7 & 0.2000&0.8000 &  0.0437 & 0.2463 & 0.6006 & 0.1094 &0.3147&0.6853 & 0.0707 & 0.1388 & 0.0001 \\
  8 & 0.0909&0.9091 &   0.0753 & 0.1813 & 0.7289 & 0.0145 &0.2039&0.7961 & 0.0358 & 0.1395 & 0.0000 \\
  9 & 0.0000&1.0000* & 0.0507 & 0.1351 & 0.8001 & 0.0141 & 0.1497&0.8503 & 0.0381 & \textbf{0.1823} & 0.0000 \\
  10 & 0.0909&0.9091 &   0.0825 & 0.1927 & 0.7089 & 0.0159 &0.2186&0.7814 & 0.0336 & 0.1115 & 0.0000 \\
  11 & 0.0588&0.9412 &   0.1063 & 0.1596 & 0.7235 & 0.0106 &0.1845&0.8155 & 0.0230 & 0.1119 & 0.0000 \\
  12 & 0.5000&0.5000 &   0.3803 & 0.3042 & 0.3060 & 0.0095 &0.4986&0.5014 & 0.0142 & 0.0143 & 0.0001 \\
   \hline
\end{tabular}}
\label{x12table}
\end{table}

\subsection{$X_{11}$ data set} 
\begin{center}\begin{figure}[!thbt] \centering
 		\includegraphics[width=0.45\linewidth]{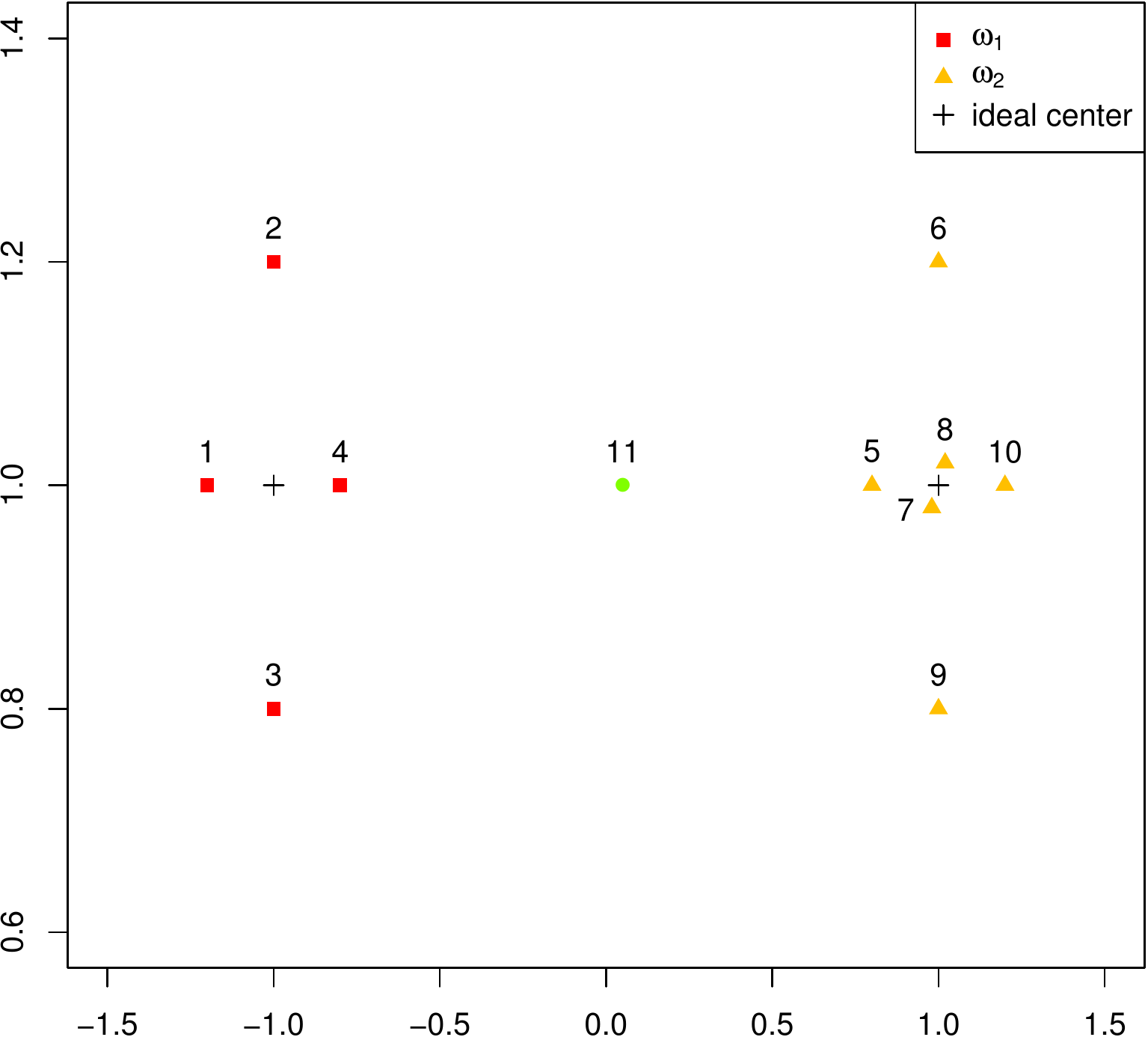}
		\hfill \caption{A simple data set of 11 objects. The ideal centers of the two clusters are located at (-1, 1) and (1, 1). The coordinates of object 11 are (0.05, 1), which is closer to the center of cluster 2.}
	\label{x11data}\end{figure} \end{center}

In this experiment, we will show the effectiveness of the application of multiple weighted prototypes using the data set displayed in Figure \ref{x11data}. The $X_{11}$ data set has two obvious clusters, one containing objects 1 to 4 and the other including objects 5 to 10. Object 11 locates slightly biased to the cluster on the right side.  It can be seen that in the left class, it is unreasonable to describe the cluster structure using any one of the four objects in the group, since no one of the four points could be viewed as a more proper representative than the other three. The clustering results by FCMdd, sECMdd, wECMdd are listed in Table~\ref{x11tabel}. The result by MECM is not listed here as it is similar to that by sECMdd.

From the table we can see that the two clustering approaches, FCMdd and sECMdd, which using a single medoid cluster to represent a cluster, partition object 11 to cluster 1 for mistake. This is resulted by the fact that both of them set  object 4 to be the center of class $\omega_1$. On the contrary,  in wECMdd, the four objects in cluster $\omega_1$ are thought to have nearly the same contribution to the class. Consequently, object 11 is clustered into $\omega_2$ correctly. FMMdd could also get the exactly accurate results as it also takes use of multiple weighted medoids. This experiment shows that the multi-prototype representation of classes could capture some complex data structure and consequently enhance the clustering performance. It is remarkable that the hard partition could be recovered from pignistic probability ($\BetP$) for credal partitions. And the results of these  experiments
reflects that  pignistic probabilities  play a similar role as fuzzy
membership.

% latex table generated in R 3.1.1 by xtable 1.7-3 package
% Sun Feb 08 22:39:53 2015
\begin{table}[ht]
\centering\caption{The clustering results of $X_{11}$ data set. The objects marked  with * are the medoids found by FCMdd and sECMdd. Values $v_{ij}, j=1,2,3$ are the weights of object $x_i$ for class $\omega_1$, $\omega_2$ and imprecise class $\omega_{12}\triangleq\{\omega_1,\omega_2\}$.}
\begin{tabular}{l|ll|ll|ll|llllllllll}
\hline
& \multicolumn{2}{c|}{FCMdd} &\multicolumn{2}{c|}{sECMdd}&\multicolumn{5}{c}{wECMdd}\\
 % \cline{2-3} \cline{5-6} \cline{8-12}
 \hline
 id & $u_{i1}$&$u_{i2}$  &$\BetP_{i1}$ &  $\BetP_{i2}$   &$\BetP_{i1}$ &  $\BetP_{i2}$  & $v_{i1}$ & $v_{i2}$ & $v_{i3}$ \\
  \hline
1 & 0.9674&0.0326 & 0.9510 & 0.0490 & 0.9620 & 0.0380 & \textbf{0.1477} & 0.0414 & 0.0018 \\
  2 & 0.9802&0.0198 & 0.9671 & 0.0329 & 0.9578 & 0.0422 & 0.1476 & 0.0433 & 0.0024 \\
  3 & 0.9802&0.0198 & 0.9667 & 0.0333 & 0.9578 & 0.0422 & 0.1476 & 0.0433 & 0.0024 \\
  4 & 1.0000&0.0000* & 1.0000 & 0.0000*  & 0.9517 & 0.0483 & 0.1475 & 0.0457 & 0.0033 \\
  5 & 0.0127&0.9873 & 0.0958 & 0.9042 & 0.0169 & 0.9831 & 0.0585 & 0.1190 & 0.0320 \\
  6 & 0.0147&0.9853 & 0.0383 & 0.9617 & 0.0145 & 0.9855 & 0.0554 & 0.1187 & 0.0223 \\
  7 & 0.0000&1.0000* & 0.0327 & 0.9673 & 0.0073 & 0.9927 & 0.0558 & \textbf{0.1447} & 0.0117 \\
  8 & 0.0010&0.9990& 0.0198 & 0.9802 & 0.0072 & 0.9928 & 0.0553 & 0.1445 & 0.0111 \\
  9 & 0.0099&0.9901& 0.5000 & 0.5000 & 0.0144 & 0.9856 & 0.0554 & 0.1187 & 0.0223 \\
  10 & 0.0121&0.9879& 0.0000 & 1.0000* & 0.0128 & 0.9872 & 0.0530 & 0.1183 & 0.0167 \\
  11 & 0.5450&0.4550& 0.5723 & 0.4277 & 0.4990 & 0.5010 & 0.0761 & 0.0625 & \textbf{0.8739} \\
   \hline
\end{tabular}
\label{x11tabel}
\end{table}
\subsection{Karate Club network}
Graph visualization is commonly used to visually model relations in many areas. For graphs such as social networks, the
prototype (center) of one group is likely to be one of the persons ({\em i.e.}
nodes in the graph) playing the leader role in the community. Moreover, a graph (network) of vertices (nodes) and edges usually describes the interactions between different agents of the complex system. The pair-wise relationships between nodes are often implied in the graph data sets. Thus medoid-based relational clustering algorithms could be directly applied. In this section we will evaluate the effectiveness of the proposed methods applied on community detection problems.

\begin{center} \begin{figure}[!thbt] \centering
		\includegraphics[width=0.45\linewidth]{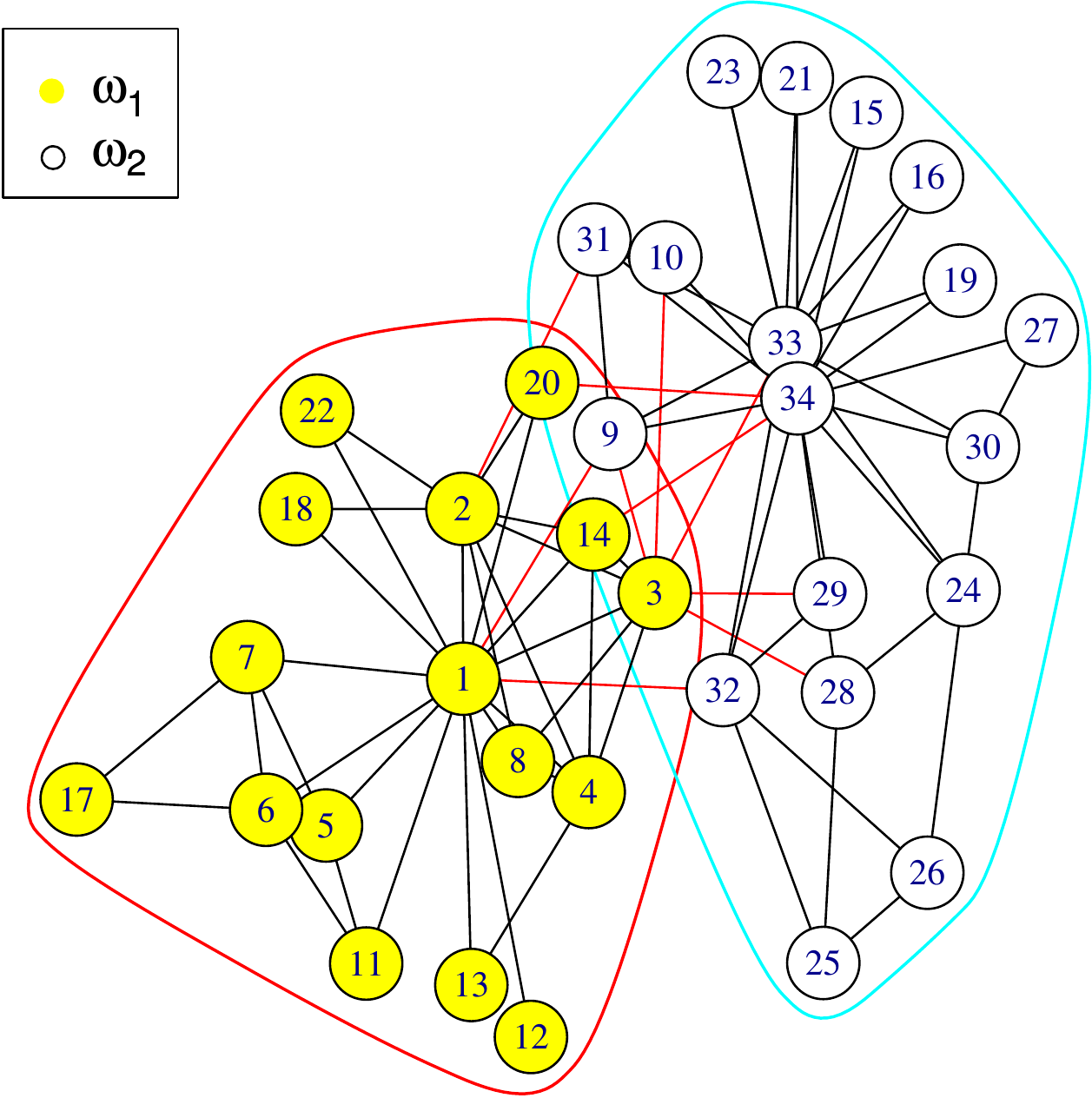}\hfill
        \includegraphics[width=0.45\linewidth]{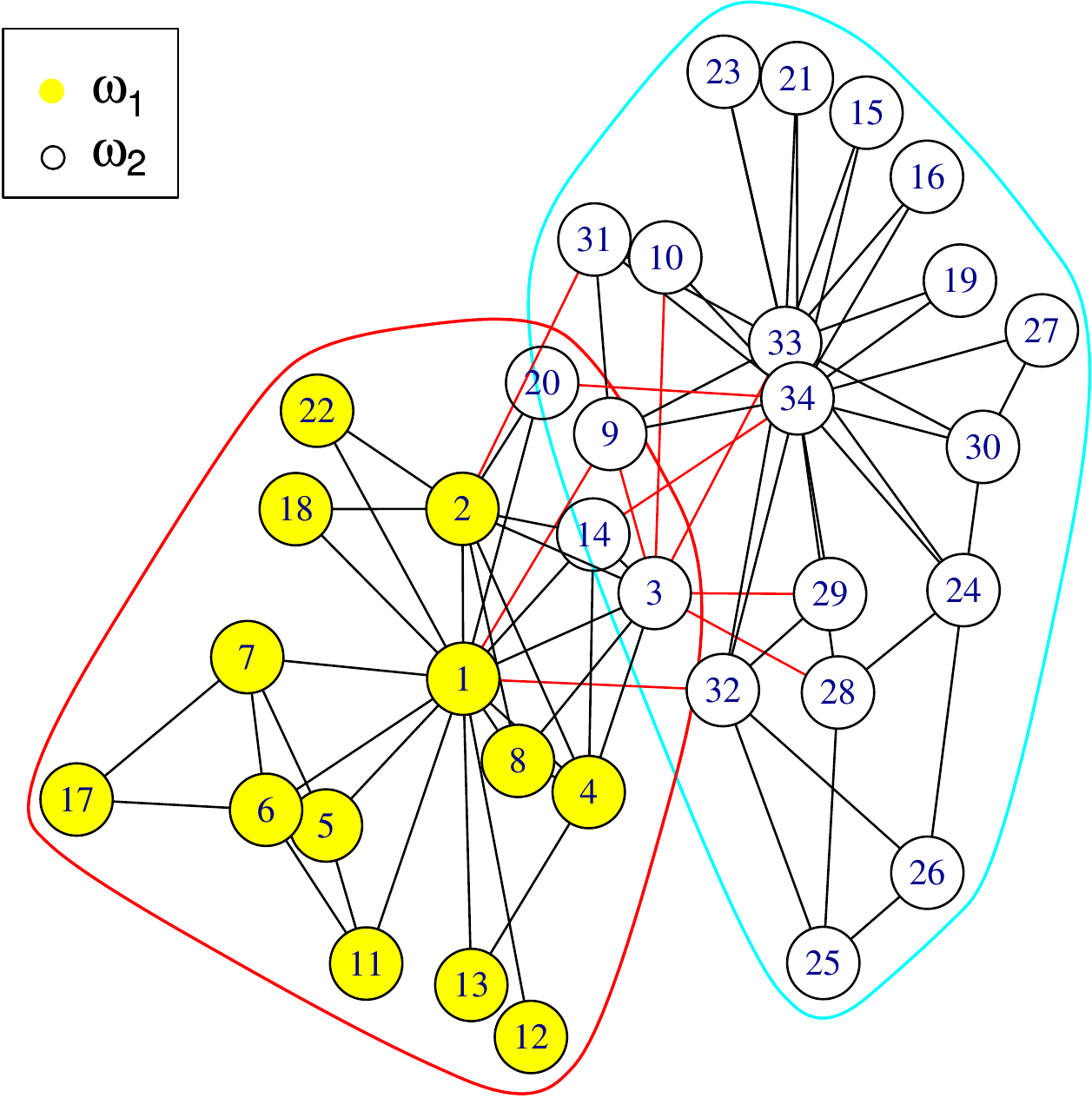} \hfill
        \parbox{.45\linewidth}{\centering\small a. Original network} \hfill
		\parbox{.45\linewidth}{\centering\small b. Results by FCMdd}
		\includegraphics[width=0.45\linewidth]{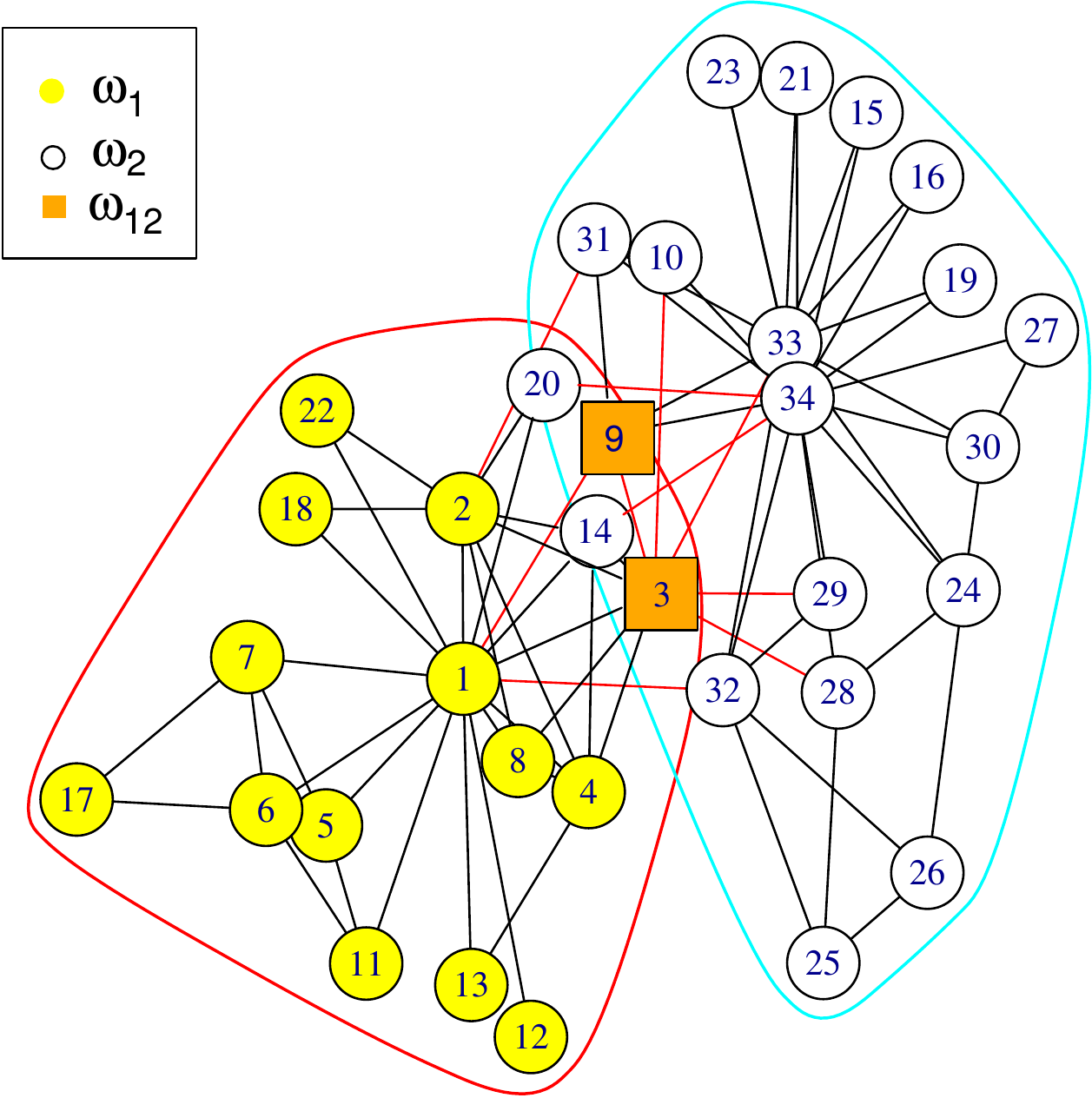}\hfill
        \includegraphics[width=0.45\linewidth]{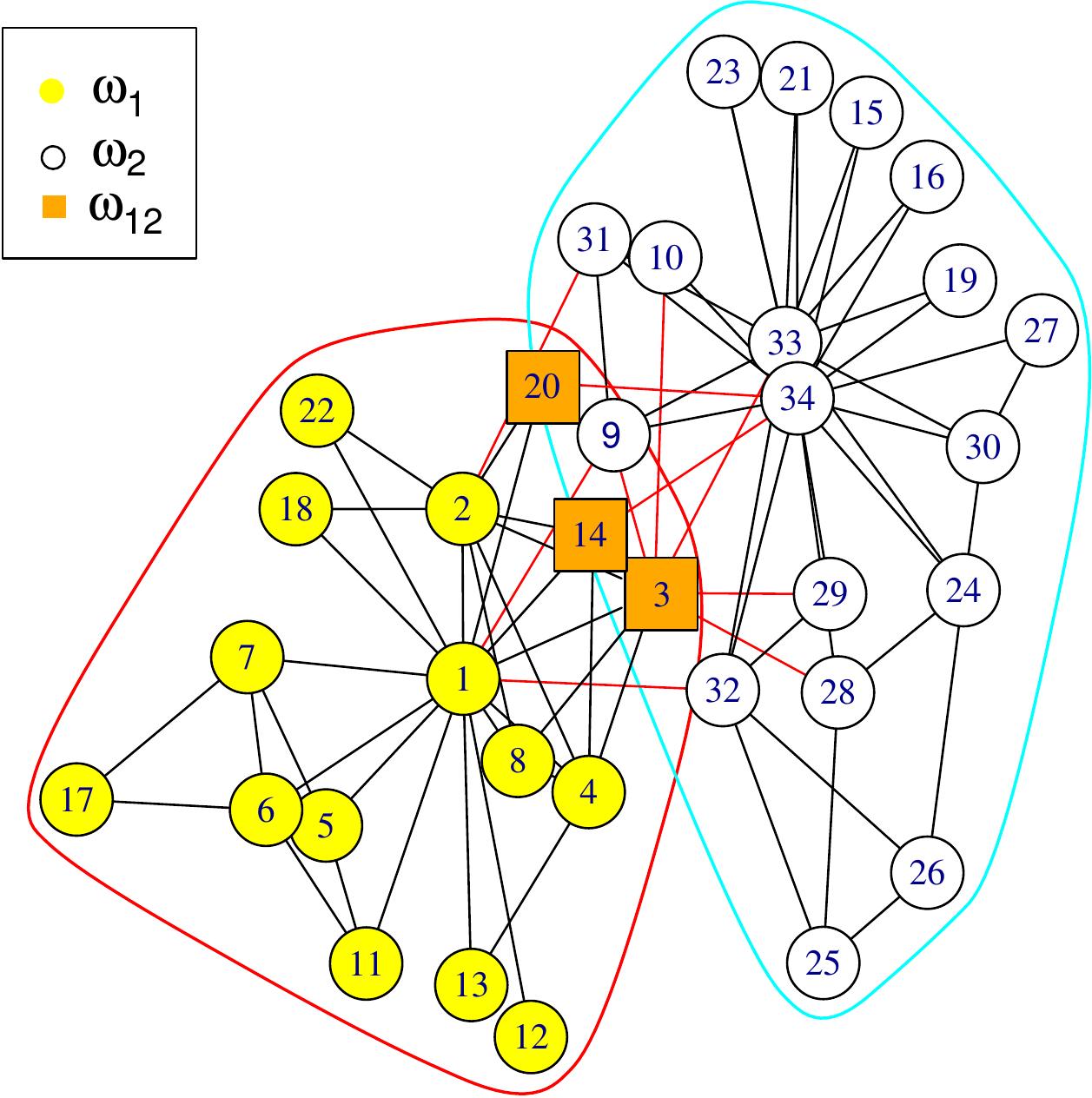} \hfill
        \parbox{.45\linewidth}{\centering\small c. Results by MECM} \hfill
		\parbox{.45\linewidth}{\centering\small d. Results by sECMdd}
\caption{The Karate Club network. The parameters of MECM are $\alpha=1.5, \beta=2, \delta=100, \eta=0.9, \gamma=0.05$. In sECMdd, $\alpha=0.05, \beta=2, \delta = 100, \eta = 1, \gamma =1$, while in FCMdd, $\beta=2$.} \label{karate} \end{figure} \end{center}

Here the widely used benchmark  in detecting community structures, ``Karate Club",  studied by Wayne Zachary is considered. The network consists of 34  nodes and 78 edges representing the friendship among the members of the club (see Figure \ref{karate}.a). During the development, a dispute arose between the club's administrator and instructor, which eventually resulted in the club  split into two smaller clubs, centered around the administrator and the instructor respectively.

There are many similarity and dissimilarity indices for networks, using local or global information of graph structure. In this experiment,
different similarity metrics will be compared first. The similarity indices considered here are listed in Table \ref{indices}~\footnote{A more detailed description could be found in the appendix.}. It is notable that the similarities by these measures range
from 0 to 1, thus they  can be converted  into dissimilarities simply by
$
  \textit{dissimilarity} = 1 - \textit{similarity}
$. The comparison results for different dissimilarity indices by FCMdd and sECMdd are shown in Tables \ref{ComIndicesFCMdd} and \ref{ComIndicessECMdd} respectively.  As we can see, for all the dissimilarity indices, for sECMdd, the value of evidential precision is higher than that of precision. This can be attributed to the introduced imprecise classes which enable us not to make hard decisions for the nodes that we are uncertain and consequently guarantee the accuracy of the specific clustering results. From the table we can also see that the performance using the dissimilarity measure based on signal prorogation  is better than those using local similarities in the application of both FCMdd and sECMdd. This reflects that global dissimilarity metric is better than the local ones for community detection. Thus in the following experiments,  only  the signal dissimilarity index is considered.
\begin{table}[ht]
\centering \caption{Different local and global similarity indices.}
\begin{tabular}{rrrrrrr}
  \hline
 Index name & Global metric& Ref.\\
  \hline
Jaccard & No &  \cite{jaccard1912distribution}\\
Pan  & No &  \cite{pan2010detecting}\\
Zhou & No &  \cite{zhou2009predicting}\\
Signal & Yes &\cite{hu2008community}\\
%Football network & \\
   \hline
\end{tabular}\label{indices}
\end{table}

\begin{table}[ht]
\centering \caption{Comparison of different similarity indices by FCMdd.}
\begin{tabular}{rrrrrrr}
  \hline
 Index & P & R & RI & EP & ER & ERI \\
  \hline
Jaccard & 0.6364 & 0.7179 & 0.6631 & 0.6364 & 0.7179 & 0.6631 \\
  Pan & 0.4866 & 1.0000 & 0.4866 & 0.4866 & 1.0000 & 0.4866 \\
  Zhou & 0.4866 & 1.0000 & 0.4866 & 0.4866 & 1.0000 & 0.4866 \\
  Signal & 0.8125 & 0.8571 & 0.8342 & 0.8125 & 0.8571 & 0.8342 \\
   \hline
\end{tabular}\label{ComIndicesFCMdd}
\end{table}

% latex table generated in R 3.1.1 by xtable 1.7-3 package
% Sun Mar 08 12:11:22 2015
\begin{table}[ht]
\centering \caption{Comparison of different similarity indices by sECMdd.}
\begin{tabular}{rrrrrrr}
  \hline
 Index & P & R & RI & EP & ER & ERI \\
  \hline
Jaccard & 0.6458 & 0.6813 & 0.6631 & 0.7277 & 0.5092 & 0.6684  \\
Pan & 0.6868 & 0.7070 & 0.7005 & 0.7214 & 0.6923 & 0.7201 \\
Zhou  & 0.6522 & 0.6593 & 0.6631 & 0.7460 & 0.3443 & 0.6239\\
Signal & 1.0000 & 1.0000 & 1.0000 & 1.0000 & 0.6190 & 0.8146 \\
   \hline
\end{tabular}\label{ComIndicessECMdd}
\end{table}

The detected community structures by different methods are displayed in Figures \ref{karate}.b -- \ref{karate}.d.  FCMdd could detect the exact community structure of all the nodes except nodes 3, 14, 20. As we can see from the figures, these three nodes have connections  with both communities.  They are partitioned into imprecise class $\omega_{12}\triangleq\{\omega_1, \omega_2\}$, which describing the uncertainty on the exact class labels of the related nodes, by the application of sECMdd.   The medoids found by FCMdd of the two specific communities are node 5 and node 29, while by sECMdd node 5 and node 33.  The uncertain nodes found by MECM are node 3 and node 9. %sECMdd seems have higher power to express the uncertainty compared with MECM. Moreover, sECMdd is more efficient than MECM in terms of executing time.
%Table \ref{pwkarate} lists the prototype weights obtained by wsECMdd-0. The nodes in each community are order by prototype weights in the table. We just display the first ten important members in every class. From the table we can see that node 1 and 12 play the center role in community $\omega_1$, while node 24 and 33 consists the two cores in community $\omega_2$. Node 9 contributes most to the overlap community $\omega_{12}$, which is a good reflection of its ``bridge" role for the two classes. Therefore, the prototype weights provide us some information about the cluster structure from another point of view, which could help us gain a better understanding of the inner structure of a class.

The results by wECMdd algorithms are similar to that by sECMdd. Table~\ref{pwkarate} lists the prototype weights obtained by FMMdd and wECMdd. The nodes in each community are ordered by prototype weights in the table. We just display the first ten important members in every class. From the weight values by FMMdd and wECMdd in the table we can get the same conclusion: nodes 1 and 12 play the center role in community $\omega_1$, while node 33 and 34 consists the two cores in community $\omega_2$. But by wECMdd more information about the overlapped structure of the network are available. As we can see from the last two columns of the table, node 9 contributes most to the overlapped community $\omega_{12}$, which is a good reflection of its ``bridge" role for the two classes. Therefore, the prototype weights provide us some information about the cluster structure from another point of view, which could help us gain a better understanding of the inner structure of a class.

%% latex table generated in R 3.1.1 by xtable 1.7-3 package
%% Mon Feb 23 11:10:25 2015
%\begin{table}[ht]
%\centering \caption{The prototype weights by wECMdd-0.}
%\resizebox{\textwidth}{!}{
%\begin{tabular}{rrrrrrrrrrrrrrrrrrrrrrrrrrrrrrrrrrr}
%  \hline
%&  & 1 & 2 & 3 & 4 & 5 & 6 & 7 & 8 & 9 & 10 \\
%  \hline
%Community $\omega_1$& Node & 12 & 1 & 18 & 22 & 13 & 4 & 8 & 2 & 5 & 11 \\
% & Weights & 0.1639 & 0.1491 & 0.1054 & 0.1054 & 0.1043 & 0.0691 & 0.0674 & 0.0651 & 0.0339 & 0.0339 \\
%  \hline
% Community $\omega_2$& Node & 24 & 33 & 30 & 15 & 16 & 19 & 21 & 23 & 34 & 27 \\
%  &Weights & 0.1131 & 0.0931 & 0.0870 & 0.0855 & 0.0855 & 0.0855 & 0.0855 & 0.0855 & 0.0653 & 0.0514 \\
%  \hline
%  Community $\omega_{12}$&Node & 9 & 10 & 32 & 29 & 20 & 3 & 31 & 25 & 14 & 28 \\
%  & Weights & 0.9943 & 0.0016 & 0.0013 & 0.0009 & 0.0008 & 0.0006 & 0.0004 & 0 & 0 & 0 \\
%   \hline
%\end{tabular}}\label{pwkarate1}
%\end{table}
% latex table generated in R 3.1.1 by xtable 1.7-3 package
% Mon Mar 23 15:09:48 2015
\begin{table}[ht]
\centering \caption{The prototype weights by FMMdd and wECMdd. Community $\omega_{12}$ denotes the imprecise community $\{\omega_1, \omega_2\}$. Only the first 10 nodes with largest weight values in each community are listed.}
\begin{tabular}{ll|ll|ll|ll|lllllllll}
  \hline
  \multicolumn{4}{l|}{FMMdd} &\multicolumn{6}{l}{wECMdd}\\
  %\cline{1-4} \cline{5-11}
  \hline
  \multicolumn{2}{l|}{Community $\omega_1$}&\multicolumn{2}{l|}{Community $\omega_2$}&\multicolumn{2}{l|}{Community $\omega_1$}
  &\multicolumn{2}{l|}{Community $\omega_2$}&\multicolumn{2}{l}{Community $\omega_{12}$}\\
  %\cline{1-2} \cline{3-4} \cline{6-7} \cline{8-9} \cline{10-11}
   \hline
    Node & Weights & Node & Weights & Node & Weights& Node & Weights& Node & Weights \\
   \hline
     1 & 0.0689 & 33 & 0.0607 &12 & 0.0707 & 33 & 0.0606 & 9 & 0.3194 \\
    12 & 0.0663 & 34 & 0.0565 & 1 & 0.0659 & 34 & 0.0562 & 3 & 0.1348 \\
    22 & 0.0590 & 28 & 0.0556 & 13 & 0.0588 & 24 & 0.0557 & 20 & 0.1254 \\
    18 & 0.0590 & 24 & 0.0551 & 18 & 0.0584 & 28 & 0.0549 & 25 & 0.0989 \\
    13 & 0.0583 & 15 & 0.0512 & 22 & 0.0584 & 15 & 0.0519 & 10 & 0.0493 \\
    2 & 0.0548 & 16 & 0.0512 & 5 & 0.0519 & 16 & 0.0519 & 32 & 0.0453 \\
    4 & 0.0544 & 19 & 0.0512 & 11 & 0.0519 & 19 & 0.0519 & 26 & 0.0429 \\
    8 & 0.0537 & 21 & 0.0512 & 4 & 0.0506 & 21 & 0.0519 & 29 & 0.0379 \\
    14 & 0.0469 & 23 & 0.0512 & 8 & 0.0503 & 23 & 0.0519 & 14 & 0.0351 \\
    5 & 0.0436 & 31 & 0.0504 & 2 & 0.0500 & 30 & 0.0509 & 31 & 0.0306 \\
   \hline
\end{tabular}\label{pwkarate}
\end{table}
\subsection{Countries data}
In this section we will test  on a  relational data set, referred as the
benchmark data set Countries Data \cite{kaufman2009finding,mei2010fuzzy}. The task is to group twelve
countries into clusters based on the pairwise relationships as given
in Table \ref{countriestable}, which is in fact the average dissimilarity scores on some dimensions of quality of life provided
subjectively by students in a political science class. Generally,
these countries are classified into three categories: Western,
Developing and Communist. % In this experiment, the original dissimilarities between countries $i$ and $j$ are scaled by $$\tau_{ij}=d_{ij}/\max \{d_{ij}\},~~ i, j= 1,2,\cdots,12. $$
The  parameters are set as $\beta=2$ for
FCMdd, and $\beta=2, \alpha = 0.95, \eta = 1, \gamma = 1$ for sECMdd. We test the performances
of FCMdd and sECMdd with two different sets of initial
representative countries: $\Delta_1=\{\text{C10: USSR; C8: Israel; C7: India}\}$ and $\Delta_2=\{\text{C6: France; C4: Cuba; C1: Belgium}\}$. The three countries in $\Delta_1$ are well separated. On the contrary,  for the countries in $\Delta_2$, Belgium
is similar to France, which makes two initial medoids of three are very close in terms of the given dissimilarities.
% latex table generated in R 3.1.1 by xtable 1.7-3 package
% Mon Feb 16 17:03:40 2015
\begin{table*}[ht]
\centering \caption{Countries data: dissimilarity matrix.}
\resizebox{\textwidth}{!}{
\begin{tabular}{llllllllllllllll}
  \hline
  & Countries & C1 & C2 & C3 & C4 & C5 & C6 & C7 & C8 & C9 & C10 & C11 & C12  \\
  \hline
1 & C1: Belgium:  & 0.00 & 5.58 & 7.00 & 7.08 & 4.83 & 2.17 & 6.42 & 3.42 & 2.50 & 6.08 & 5.25 & 4.75 \\
  2 & C2: Brazil & 5.58 & 0.00 & 6.50 & 7.00 & 5.08 & 5.75 & 5.00 & 5.50 & 4.92 & 6.67 & 6.83 & 3.00 \\
  3 & C3: China & 7.00 & 6.50 & 0.00 & 3.83 & 8.17 & 6.67 & 5.58 & 6.42 & 6.25 & 4.25 & 4.50 & 6.08 \\
  4 & C4: Cuba & 7.08 & 7.00 & 3.83 & 0.00 & 5.83 & 6.92 & 6.00 & 6.42 & 7.33 & 2.67 & 3.75 & 6.67 \\
  5 & C5: Egypt & 4.83 & 5.08 & 8.17 & 5.83 & 0.00 & 4.92 & 4.67 & 5.00 & 4.50 & 6.00 & 5.75 & 5.00 \\
  6 & C6: France & 2.17 & 5.75 & 6.67 & 6.92 & 4.92 & 0.00 & 6.42 & 3.92 & 2.25 & 6.17 & 5.42 & 5.58 \\
  7 & C7: India & 6.42 & 5.00 & 5.58 & 6.00 & 4.67 & 6.42 & 0.00 & 6.17 & 6.33 & 6.17 & 6.08 & 4.83 \\
  8 & C8: Israel & 3.42 & 5.50 & 6.42 & 6.42 & 5.00 & 3.92 & 6.17 & 0.00 & 2.75 & 6.92 & 5.83 & 6.17 \\
  9 & C9: USA & 2.50 & 4.92 & 6.25 & 7.33 & 4.50 & 2.25 & 6.33 & 2.75 & 0.00 & 6.17 & 6.67 & 5.67 \\
  10 & C10: USSR & 6.08 & 6.67 & 4.25 & 2.67 & 6.00 & 6.17 & 6.17 & 6.92 & 6.17 & 0.00 & 3.67 & 6.50 \\
  11 & C11: Yugoslavia & 5.25 & 6.83 & 4.50 & 3.75 & 5.75 & 5.42 & 6.08 & 5.83 & 6.67 & 3.67 & 0.00 & 6.92 \\
  12 & C12: Zaire & 4.75 & 3.00 & 6.08 & 6.67 & 5.00 & 5.58 & 4.83 & 6.17 & 5.67 & 6.50 & 6.92 & 0.00 \\
   \hline
\end{tabular}}\label{countriestable}
\end{table*}
%\fi

The results of FCMdd and sECMdd are given in
Table \ref{couFCMdd} and Table \ref{cousECMdd} respectively.  It can be seen that FCMdd is very sensitive to
initializations.  When the initial prototypes are well set (the case of $\Delta_1$), the obtained partition is reasonable. However, the clustering results become worse when the initial medoids are not ideal (the case of $\Delta_2$). In fact two of the three medoids are not changed during the update process of FCMdd when using initial prototype set $\Delta_2$. This example illustrates that FCMdd is quite easy to be stuck in a
local minimum. For sECMdd, the credal partitions are the same with
different initializations. The pignistic probabilities are also displayed in Table \ref{cousECMdd}, which could be regarded as membership values in fuzzy partitions. The country Egypt is clustered into imprecise class \{1, 2\}, which indicating that Egypt is not so well belonging  to
Developing or Western alone, but belongs to both categories. This result is consistent with the fact shown from the dissimilarity matrix: Egypt is similar to both USA and India, but has the largest dissimilarity to China. The results by wECMdd and MECM algorithms are not displayed here, as they product the same clustering result with sECMdd. From this experiment we could conclude that ECMdd is more robust to the initializations than FCMdd.
% latex table generated in R 3.1.1 by xtable 1.7-3 package
% Sun Mar 08 21:51:04 2015
\begin{table*}[ht]
\centering \caption{Clustering results of FCMdd for countries data. The prototype (medoid) of each class is marked with *.}
\resizebox{\textwidth}{!}{
\begin{tabular}{llllllllllllll}
\hline
& & \multicolumn{5}{l}{FCMdd with $\Delta_1$} & &\multicolumn{5}{l}{FCMdd with $\Delta_2$}\\
\cline{3-7} \cline{9-13}
 & Countries  & $u_{i1}$ & $u_{i2}$ & $u_{i3}$ & Label & Medoids &  & $u_{i1}$ & $u_{i2}$ & $u_{i3}$ & Label & Medoids \\
  \hline
1 &        C1:    Belgium &    0.4773  &    0.2543  &    0.2685  &          1 &          -& &    1.0000  &    0.0000  &    0.0000  &          1 &          * \\

         2 &        C6:     France &    0.4453  &    0.2719  &    0.2829  &          1 &          -& &    0.0000  &    1.0000  &    0.0000  &          2 &          * \\

         3 &        C8:     Israel &    1.0000  &    0.0000  &    0.0000  &          1 &          * &  &  0.4158  &    0.3627  &    0.2215  &          1 &          - \\

         4 &        C9:        USA &    0.5319  &    0.2311  &    0.2371  &          1 &          -& &    0.4078  &    0.4531  &    0.1391  &          2 &          - \\ \\

         5 &        C3:      China &    0.2731  &    0.3143  &    0.4126  &          3 &          -& &    0.2579  &    0.2707  &    0.4714  &          3 &          - \\

         6 &        C4:       Cuba &    0.2235  &    0.2391  &    0.5374  &          3 &          -& &    0.0000  &    0.0000  &    1.0000  &          3 &          * \\

         7 &       C10:       USSR &    0.0000  &    0.0000  &    1.0000  &          3 &          * &&    0.2346  &    0.2312  &    0.5342  &          3 &          - \\

         8 &       C11: Yugoslavia &    0.2819  &    0.2703  &    0.4478  &          3 &          -& &    0.2969  &    0.2875  &    0.4156  &          3 &          - \\ \\

         9 &        C2:     Brazil &    0.3419  &    0.3761  &    0.2820  &          2 &          -& &    0.3613  &    0.3506  &    0.2880  &          1 &          - \\

        10 &        C5:      Egypt &    0.3444  &    0.3687  &    0.2870  &          2 &          -& &    0.3558  &    0.3493  &    0.2948  &          1 &          - \\

        11 &        C7:      India &    0.0000  &    1.0000  &    0.0000  &          2 &          *& &    0.3257  &    0.3257  &    0.3485  &          3 &          - \\

        12 &       C12:      Zaire &    0.3099  &    0.3959  &    0.2942  &          2 &          -& &    0.3901  &    0.3321  &    0.2778  &          1 &          - \\

   \hline
\end{tabular}}\label{couFCMdd}
\end{table*}
\begin{table*}[ht]
\centering \caption{Clustering results of sECMdd for countries data. The prototype (medoid) of each class is marked with *. The Label \{1, 2\} represents the imprecise class expressing the uncertainty on class 1 and class 2.}
\resizebox{\textwidth}{!}{
\begin{tabular}{llllllllllllllll}
\hline
& & \multicolumn{5}{l}{sECMdd with $\Delta_1$} & &\multicolumn{5}{l}{sECMdd with $\Delta_2$}\\
\cline{3-7} \cline{9-13}
 & Countries  & $\BetP_{i1}$ & $\BetP_{i2}$ & $\BetP_{i3}$ & Label & Medoids &  & $\BetP_{i1}$ & $\BetP_{i2}$ & $\BetP_{i3}$ & Label & Medoids \\
  \hline
1 & C1: Belgium & 1.0000 & 0.0000 & 0.0000 &  1 & * & & 1.0000 & 0.0000 & 0.0000 & 1 & * \\
  2 & C6: France & 0.4932 & 0.2633 & 0.2435 & 1 & - && 0.5149 & 0.2555 & 0.2297 & 1 & - \\
  3 & C8: Israel & 0.4144 & 0.3119 & 0.2738 & 1 & - && 0.4231 & 0.3051 & 0.2719 & 1 & - \\
  4 & C9: USA & 0.4503 & 0.2994 & 0.2503    & 1 & - && 0.4684 & 0.2920 & 0.2396 & 1 & - \\ \\
  5 & C3: China& 0.2323 & 0.2294 & 0.5383 & 3   & * && 0.0000 & 0.0000 & 1.0000 & 3 & * \\
  6 & C4: Cuba & 0.2778 & 0.2636 & 0.4586 & 3   & - && 0.2899 & 0.2794 & 0.4307 & 3 & - \\
  7 & C10: USSR& 0.2509 & 0.2260 & 0.5231 & 3  & -& & 0.3167 & 0.2849 & 0.3984 & 3 & - \\
  8 & C11: Yugoslavia & 0.3478 & 0.2488 & 0.4034 & 3 & -& & 0.3579 & 0.2526 & 0.3895 & 3 & - \\
 \\  9 & C2: Brazil & 0.0000 & 1.0000 & 0.0000 & 2 & *& & 0.0000 & 1.0000 & 0.0000 & 2 & * \\
  10 & C5: Egypt & 0.3755 & 0.3686 & 0.2558 &\{1, 2\} & -& & 0.3845 & 0.3777 & 0.2378 & \{1, 2\} & - \\
  11 & C7: India& 0.3125 & 0.3650 & 0.3226 & 2 & - && 0.2787 & 0.3740 & 0.3473 & 2 & - \\
  12 &  C12: Zaire & 0.3081 & 0.4336 & 0.2583 & 2 & -& & 0.3068 & 0.4312 & 0.2619 & 2 & - \\
   \hline
\end{tabular}}\label{cousECMdd}
\end{table*}

\subsection{UCI data sets}
Finally the clustering performance of different methods will be compared on eight benchmark UCI  data sets \cite{Lichman:2013}
summarized in Table \ref{ucidatalist}.
Euclidean distance is used  as the dissimilarity measure for the object data sets, and the Signal dissimilarity is adopted for the graph data sets.

\begin{table}[ht]
\centering\caption{A summary of eight UCI data sets.}
\begin{tabular}{lllllll}
  \hline
  Data set & No. of objects & No. of cluster & Category\\
  \hline
Iris& 150 & 3 & object data\\
Cat cortex & 65 & 4 & relational data\\
Protein & 213 & 4& relational data\\
American football & 115 & 12 & graph  data \\
Banknote &  1372 & 2 & object data\\
Segment &  2100 & 19 & object data\\
Digits  & 1797 & 10 & object data\\
Yeast  & 1484 & 10 & object data\\
   \hline
\end{tabular}\label{ucidatalist}
\end{table}
%: ``Iris flower", ``Cat cortex", ``Protein" data sets, and American college football network.
%The first is object data while
%the other three are relational data sets. The given information for ``cat cortex" and ``proteins" data is pair-wise relationship values. For the former it is a matrix of connection strengths between 65 cortical areas of the cat brain, while for the latter is  a dissimilarity matrix measuring
%the structural proximity of 213 proteins sequences. College football  is a network of American football games
%between Division IA colleges during regular season fall 2000.  The vertices in the network represent 115 teams, while the links denote 613
%regular-season games between the two teams they connect. The teams are divided
%into 12 conferences containing around 8--12 teams each and generally games are
%more frequent between members from the same conference than between those
%from different conferences.

Same as ECM, the number of parameters to be optimized in ECMdd is exponential and depends on the number of clusters \citep{masson2008ecm}. For the
number of classes larger than
10, calculations are not tractable. But we can only consider  a subclass with a
limited number of focal sets \citep{masson2008ecm}. Here we
constrain the focal sets to be composed of at most two classes (except
$\Omega$). The evaluation results are listed in Tables \ref{uciiris}--\ref{uciyeast}.

It can be seen that generally wECMdd works better than the other approaches
on all of the data sets, except for Iris data set where sECMdd works best.
This may be explained by the fact that, Iris is a small data set and each class can be well represented by one prototype. wECMdd has better
performance for the other complex data sets, since the single prototype seems not enough to capture a cluster in these cases, whereas the cluster can
be properly characterized by the multiple prototypes as done in wECMdd.  From the tables we can see that the EP values for credal partitions
by sECMdd and wECMdd are significantly higher than those for hard or fuzzy partitions, which indicates  the accuracy of specific decisions.
Consequently it will avoid the risk of misclassification by the concept of imprecise decisions.
%As we mentioned, the increase  of precision will cause the decrease of recall.

The value of ER describes the fraction of instances grouped into an
identical specific cluster out of those relevant pairs in the ground-truth. If the objects are located in the overlap, they are likely to be clustered
into imprecise classes by ECMdd. This will increase the value of EP. However,
 as few objects are partitioned into specific classes, the value of ER will
decrease. That's why for Iris data set the partitional result by wECMdd has the highest EP value
following with a low ER value. The value of ERI can be regarded as a compromise between EP and ER, and it is
an integration of EP and ER. As can be seen from the results, ECMdd performs best
in terms of ERI for  most of the data sets.
In practice, one can adjust the value of parameter $\alpha$ to get partitions with different definition.
The elapsed time for every clustering algorithm is illustrated in the last
column of each table. In terms of computational time, as excepted, the evidential clustering algorithms take more time than
hard or fuzzy clustering. But sECMdd and wECMdd are much faster than MECM. wECMdd is less time-consuming than sECMdd.

\noindent\textbf{Remark 4:} It should be noted that there is no imprecise class obtained by PAM, FCMdd, and FMMdd.
In this case,  the values of EP, ER, and ERI for the clustering results
are equal to P, R, and RI respectively. That's why the increase of EP does
not cause the decrease ER significantly. However, there are some  imprecise classes provided by MECM and ECMdd clustering algorithms.
If EP is high, it indicates that there are quite a number of objects that we could not make
specific decisions and have to be clustered into imprecise classes to avoid misclassification. Thus
there will be few number of objects clustered into specific classes. Consequently
the value of ER will be declined.
%As we mentioned, the value of $\alpha$ can be adjusted to balance between EP and ER based on the requirement.
%ERI can be regarded as an integrated index considering both EP and  ER.

%The results on other data sets further proves the effectiveness of multiple prototype representativeness of classes.

%\begin{center} \begin{figure}[!thbt] \centering
%		\includegraphics[width=0.45\linewidth]{uci_p.pdf}\hfill
%        \includegraphics[width=0.45\linewidth]{uci_recall.pdf} \hfill
%        \parbox{.45\linewidth}{\centering\small a. Precision} \hfill
%		\parbox{.45\linewidth}{\centering\small b. Recall}
%		\includegraphics[width=0.45\linewidth]{uci_ri.pdf}\hfill
%        \parbox{.6\linewidth}{\centering\small c. RI}
%\caption{The clustering results for  UCI data sets.} \label{uci4} \end{figure} \end{center}
% latex table generated in R 3.1.1 by xtable 1.7-3 package
% Sun Mar 22 12:13:28 2015
\begin{table}[htp]
\centering\caption{The clustering results on Iris data set.}
\begin{tabular}{rrrrrrrrrrrr}
  \hline
 & P & R & RI & EP & ER & ERI & Elapsed Time (s)\\
  \hline
PAM & 0.8077 & \bf{0.8571} & 0.8859 & 0.8077 & 0.8571 & 0.8859 &0.0140\\
  FCMdd & 0.7965 & 0.8520 & 0.8797 & 0.7965 & 0.8520 & 0.8797 & 0.0160\\
  FMMdd & 0.8329 & 0.8411 & 0.8923 & 0.8329 & \bf{0.8411} & 0.8923 &0.0560 \\
  MECM & 0.8347 & 0.8384 & 0.8923 & 0.9454 & 0.7064 & 0.8900&73.3300 \\
  sECMdd & \bf{0.8359} & 0.8471 & \bf{0.8950} & 0.9347 & 0.7328 & \bf{0.8953}&0.2500 \\
  wECMdd & 0.8305 & 0.8335 & 0.8893 & \bf{0.9742} & 0.4827 & 0.8257 &0.2000\\
   \hline
\end{tabular}\label{uciiris}
\end{table}
% latex table generated in R 3.1.1 by xtable 1.7-3 package
% Sun Mar 22 12:14:57 2015
\begin{table}[htp]
\centering\caption{The clustering results on Proteins data set.}
\begin{tabular}{rrrrrrrrrrrrrrrrrrrr}
  \hline
 & P & R & RI & EP & ER & ERI & Elapsed Time (s)\\
  \hline
PAM & 0.7023 & 0.8246 & 0.8492 & 0.7023 & 0.8246 & 0.8492&0.0230 \\
  FCMdd & 0.6405 & 0.8353 & 0.8181 & 0.6405 & \bf{0.8353} & 0.8181&0.0200 \\
  FMMdd & 0.6586 & 0.7735 & 0.8198 & 0.6586 & 0.7735 & 0.8198&0.1760  \\
  MECM & 0.6734 & 0.8250 & 0.8348 & 0.8530 & 0.5946 & 0.8542 & 220.7700\\
  sECMdd & 0.6534 & 0.8150 & 0.7848 & \bf{0.8630} & 0.5146 & 0.8642 &0.8100\\
  wECMdd & \bf{0.7449} & \bf{0.8594} & \bf{0.8751} & 0.8609 & 0.7527 & \bf{0.8940}&0.4700 \\
   \hline
\end{tabular}\label{uciproteins}
\end{table}
% latex table generated in R 3.1.1 by xtable 1.7-3 package
% Sun Mar 22 12:16:26 2015
\begin{table}[htp]
\centering\caption{The clustering results on Cats data set.}
\begin{tabular}{rrrrrrrrrrrrrrrrrrrrr}
  \hline
 & P & R & RI & EP & ER & ERI & Elapsed Time (s)\\
  \hline
PAM & 0.6883 & 0.6897 & 0.8438 & 0.6883 & 0.6897 & 0.8438 &0.0040 \\
  FCMdd & 0.6036 & 0.5747 & 0.7986 & 0.6036 & 0.5747 & 0.7986 &0.0220\\
  FMMdd & 0.4706 & 0.6130 & 0.7298 & 0.4706 & 0.6130 & 0.7298 &0.0090 \\
  MECM & 0.7269 & 0.7088 & 0.8601 & 0.9412 & 0.3065 & 0.8212 & 8.8000\\
  sECMdd & 0.7569 & 0.7288 & 0.8801 & \bf{0.9512} & 0.2865 & 0.8312 &0.1700\\
  wECMdd & \bf{0.8526} & \bf{0.8755} & \bf{0.9308} & 0.8774 & \bf{0.8908} & \bf{0.9413} &0.1400\\
   \hline
\end{tabular}\label{ucicats}
\end{table}
% latex table generated in R 3.1.1 by xtable 1.7-3 package
% Sun Mar 22 12:17:23 2015
\begin{table}[htp]
\centering\caption{The clustering results on American football network.}
\begin{tabular}{rrrrrrrrrrrrrrrrrrrrrr}
  \hline
 & P & R & RI & EP & ER & ERI & Elapsed Time (s)\\
  \hline
PAM & 0.8649 & 0.9178 & 0.9820 & 0.8649 & \bf{0.9178} & 0.9820& 0.0430 \\
  FCMdd & 0.8649 & 0.9178 & 0.9820 & 0.8649 & 0.9178 & 0.9820 &0.0200 \\
  FMMdd & 0.8590 & 0.9082 & 0.9808 & 0.8590 & 0.9082 & 0.9808& 0.0710 \\
  MECM & 0.8232 & 0.9082 & 0.9771 & 0.9303 & 0.8681 & \bf{0.9843}&154.9300 \\
  sECMdd & 0.4166 & 0.6826 & 0.8984 & 0.7696 & 0.3384 & 0.9391&19.4700 \\
  wECMdd & \bf{0.8924} & \bf{0.9197} & \bf{0.9847} & \bf{0.9735} & 0.5621 & 0.9638&18.2100 \\
   \hline
\end{tabular}\label{ucifootball}
\end{table}

\begin{table}[htp]
\centering\caption{The clustering results on Banknote authentication data set.}
\begin{tabular}{rrrrrrrrrrrrrrrrrrrrrr}
  \hline
 & P & R & RI & EP & ER & ERI & Elapsed Time (s)\\
  \hline
  PAM & 0.5252 & 0.5851 & 0.5226 & 0.5252 & 0.5851 & 0.5226 & 0.7561 \\
  FCMdd & 0.5252 & 0.5851 & 0.5226 & 0.5252 & \textbf{0.5851} & 0.5226 & 0.8350 \\
  FMMdd & 0.5225 & 0.5302 & 0.5173 & 0.5225 & 0.5302 & 0.5173 & 5.9381 \\
  MECM & 0.5201 & 0.5618 & 0.5265 & 0.5553 & 0.4078 & 0.5353  & 50.0890 \\
  sECMdd & 0.5211 & \textbf{0.6334} & 0.5202 & 0.5191 & 0.5256 & 0.5138 & 8.2880 \\
  wECMdd & \textbf{0.5259} & 0.5645 & \textbf{0.5793} & \textbf{0.5713} & 0.4808 & \textbf{0.5797} & 7.1500 \\
   \hline
\end{tabular}\label{ucibank}
\end{table}

\begin{table}[htp]
\centering\caption{The clustering results on Segment data set.}
\begin{tabular}{rrrrrrrrrrrrrrrrrrrrrrrr}
  \hline
 & P & R & RI & EP & ER & ERI & Elapsed Time (s)\\
  \hline
PAM  & 0.4131 & 0.4910 & 0.8281 & 0.4131 & 0.4910 & 0.8281 & 7.8250\\
  FCMdd & 0.4380 & 0.5683 & 0.8246 & 0.4380  & 0.5683 & 0.8346  &8.9900\\
  FMMdd & 0.5186 & 0.8043 & 0.5626 & 0.5186 & \textbf{0.8043} & 0.5626  &107.3040\\
  MECM   &  0.5164 & 0.7744 & 0.6160 & 0.6764 & 0.5444 & 0.7160 &765.8800\\
  sECMdd  & 0.5040 & 0.7738 & 0.6065 & 0.7040 & 0.4738 & 0.7255 &351.0800\\
  wECMdd & \textbf{0.5433} & \textbf{0.8350} & \textbf{0.8455} & \textbf{0.7584} & 0.4856 & \textbf{0.8582} &308.3100\\
   \hline
\end{tabular}\label{uciSegment}
\end{table}

\begin{table}[htp]
\centering\caption{The clustering results on Digits data set.}
\begin{tabular}{rrrrrrrrrrrrrrrrrrrrrrr}
  \hline
 & P & R & RI & EP & ER & ERI & Elapsed Time (s)\\
  \hline
PAM  & 0.5928 & 0.6351 & 0.8203 & 0.5928 & 0.6351 & \textbf{0.8203} & 6.3638  \\
  FCMdd & 0.5096 & 0.5753 & 0.8026 & 0.5096 & 0.5753 & 0.8026 & 4.1913  \\
      FMMdd & 0.6542 & 0.5941 & 0.7861 & 0.6542 & 0.5941 & 0.7861 & 25.7530\\
  MECM   & 0.6148 & 0.5685 & 0.7772 & 0.8137 & 0.7268 & 0.6126 & 524.2380\\
  sECMdd & 0.7201 & 0.5920 & 0.7566 &  0.8048& \textbf{0.7630} & 0.6005 & 215.5220   \\
  wECMdd & \textbf{0.7250} & \textbf{0.6645} & \textbf{0.8232} & \textbf{0.8211} & 0.5911 & 0.8141 & 206.5590\\
   \hline
\end{tabular}\label{ucidigits}
\end{table}

\begin{table}[htp]
\centering\caption{The clustering results on Yeast data set.}
\begin{tabular}{rrrrrrrrrrrrrrrrrrrrrrr}
  \hline
 & P & R & RI & EP & ER & ERI & Elapsed Time (s)\\
  \hline
PAM & 0.5229 & 0.4848 & 0.7322 & 0.5229 & 0.4848 & 0.7322 & 4.6414 \\
  FCMdd & 0.5939 & 0.5151 & 0.7515 & 0.5939 & 0.5151 & 0.7515 & 4.7177 \\
  FMMdd & 0.5938 & \textbf{0.5568} & 0.6345 & 0.5938 & 0.5568 & 0.6345 & 12.7288 \\
  MECM & 0.3991 & 0.4098 & 0.6829 & 0.5723 & 0.5601 & 0.7149  & 212.6400 \\
  sECMdd & 0.4123 & 0.4698 & 0.7050 & 0.6393 & 0.5369 & 0.7273 & 155.5300 \\
  wECMdd & \textbf{0.6329}& 0.5065 & \textbf{0.7712} & \textbf{0.7041} & \textbf{0.6544} & \textbf{0.7917} & 134.8950 \\
   \hline
\end{tabular}\label{uciyeast}
\end{table}
Presented results allow us to sum up the characteristics of the proposed ECMdd clustering approaches (including sECMdd and wECMdd). Firstly, credal partitions provided by all the ECMdd algorithms could recover
the information of crisp and fuzzy partitions.  Secondly, ECMdd is more robust to the outliers and the initialization than FCMdd. Thirdly, the imprecise classes by credal partitions  enable us to make soft decisions for uncertain objects and could avoid the risk of misclassifications. Moreover, wECMdd performs best  generally due to the efficient class representativeness strategy. Lastly, the prototype weights provided by wECMdd algorithms are useful for our better understanding of cluster structure in real applications.

%\subsection{Discussion}
Although the computational time of wECMdd is significantly reduced compared with that of
MECM, the proposed algorithm is still of high complexity compared with hard or fuzzy clustering
algorithms such as PAM, FCMdd, and FMMdd.  However, here we discuss some possible solutions  to  further
reduce the complexity. Firstly, the number of parameters to be
optimized is exponential and depends on the number of clusters \cite{masson2008ecm}. For the number of classes
larger than 10, calculations are not tractable. But we can consider only a subclass with a limited
number of focal sets \cite{masson2008ecm}. For instance, we could constrain the focal sets to be composed of at most
two classes (except $\Omega$). Secondly, for the data sets with millions of data, some hierarchical clustering algorithms
 could be evoked as a first step to merge some objects into small groups. After that we can apply ECMdd to
the ``coarsened" data set. But how to define the dissimilarities in the new data set should be
studied. Lastly we emphasize that ECMdd is designed to detect the imprecise class structures. For the
large-scale data set, some classes may be well separated while some others may overlap. In real applications,
it is not necessary to apply ECMdd on the whole data set, but only on the special parts which may have large overlap.

\section{Conclusion}
In this contribution, the evidential $c$-medoids  clustering is proposed as a new medoid-based
clustering algorithm. Two versions of ECMdd algorithms are presented. One uses a single medoid to represent each class, while the other adopts the multiple
weighted medoids. The proposed approaches are some extensions of  crisp $c$-medoids and
fuzzy $c$-medoids on the framework of belief function theory. The experimental results illustrate the advantages of credal partitions by sECMdd and wECMdd.  Moreover, the way of using prototype weights to represent a cluster enables wECMdd to capture the various types of cluster structures more precisely and completely hence improves the quality of the detected classes. Furthermore, more detailed information on the discovered clusters may be obtained with the help of  prototype weights.

As we analyzed in the paper, assigning weights of a class to all the patterns  seems not rational since objects in other clusters make little  contribution. Thus it is better to set the number of possible objects holding positive weights differently for each class. But how to determine the optimal number of prototypes is a key problem and we will study this in our future work.  The relational descriptions of a data set may be given by multiple dissimilarity matrices.  Thus another interesting work  aiming to obtain a collaborative role of the different dissimilarity matrices
to get a final consensus partition will also be investigated in the future.

\section*{Appendix. The similarity indices for graphs.}
Here we give a detailed description of the similarity measures for graphs discussed in this paper. Let $G(V,E)$ be an undirected network, where $V$ is the set of  $N$ nodes and $E$ is the sets of $m$ edges. Let $\bm{A}=(a_{ij})_{N\times N}$ denote the adjacency matrix, where $a_{ij}=1$ represents that there is an edge between node $i$ and $j$.

\begin{description}
  \item [(1)] Jaccard Index. This index was proposed by Jaccard over a hundred years ago, and is defined as
   \begin{equation}\label{Jaccard}
            s^{\text{J}}(x,y)=\frac{|N(x) \cap N(y)|}{|N(x) \cup N(y)|},
        \end{equation}
        where $N(x)=\{w\in V\setminus x: a(w,x)=1\}$ denotes the set of vertices that are adjacent to $x$.
  \item [(2)] Zhou's Index. \citet{zhou2009predicting} also proposed a new similarity metric which is motivated by the resource allocation process:
       \begin{equation}
         s^{\text{Z}}(x,y)=\sum_{z \in N(x) \cap N(y)} \frac{1}{d(z)},
       \end{equation}
where $d(z)$ is the degree of node $z$.

 \item [(3)] Pan's Index.
\citet{pan2010detecting} pointed out that the similarity measure proposed by \citet{zhou2009predicting} may bring about inaccurate results for community detection on the networks as the metric can not differentiate the
tightness relation between a pair of nodes whether they are connected directly or indirectly. In order to overcome this defect, in his presented new measure the similarity between unconnected pair is simply set to be 0:
\begin{equation}
  S^P(x,y)=\begin{cases}
   \sum\limits_{z \in N(x) \cap N(y)} \frac{1}{d(z)}, & \text{if}~ x, y ~\text{are connected},\\
   0 & \text{otherwise}.
  \end{cases}
\end{equation}

\item [(4)] Signal similarity. A similarity measure considering the global graph structure is put forward by \citet{hu2008community} based on signaling propagation in the network.  For a network with $N$ nodes, every node is viewed as an excitable system which can send, receive, and record signals. Initially, a node is selected as the source of signal. Then
the source node sends a signal to its neighbors and itself first.
Afterwards, the nodes with signals can also send signals to
their neighbors and themselves.  After a
certain $T$ time steps, the amount distribution of signals over
the nodes could be viewed as the influence of the source
node on the whole network. Naturally, compared with nodes in other communities, the nodes of the same community have more similar
influence on the whole network. Therefore, similarities between nodes can be obtained by calculating the differences between the amount of signals they have received.
\end{description}
\section*{Acknowledgements}
This work was supported by the National
Natural Science Foundation of China (Nos.61135001, 61403310). The study of the
first author in France was supported by the China Scholarship Council.
\bibliographystyle{elsarticle-num-names}
\addcontentsline{toc}{section}{\refname}\bibliography{paperlist}

\end{document}